\pdfoutput=1

\documentclass[10pt,twoside,a4paper,fleqn]{report}
\usepackage{mathtools}
\usepackage{amsthm}
\newtheorem{remark}{Remark}
\usepackage{parskip}
\usepackage{hyperref}
\usepackage{graphicx}
\usepackage{caption}
\usepackage{algorithm}
\usepackage{algorithmicx}
\usepackage{algpseudocode}
\usepackage{subcaption}
\graphicspath{{img/}}
\usepackage[english,mt]{ethidsc} 
							
\title{Distributed Estimation, Control and Coordination of Quadcopter Swarm Robots}

\studentA{Zheng Jia}
\ethidA{13-947-254}
\semesterA{6}
\emailA{zjia@student.ethz.ch}

\supervision{Michael Hamer\\ Prof. Dr. Raffaello D'Andrea}
\date{March 29, 2017}

\identification{IDSC-RD-MDH-07} 		

\infopage
\declaration

\begin{document}

\maketitle 							


\pagenumbering{roman} 				






 \setcounter{tocdepth}{2}
 \tableofcontents

















\pagestyle{fancy}               	
\pagenumbering{arabic}				

\newcommand{\neighbor}{\mathcal{N}} 
\newcommand{\onevect}{\mathbf{1}}
\newcommand{\wij}{w_{ij}}
\newcommand{\gii}{g_{ii}}
\newcommand{\krp}{k_{rp}}
\newcommand{\krv}{k_{rv}}
\newcommand{\kap}{k_{ap}}
\newcommand{\kav}{k_{av}}
\newcommand{\kron}{\otimes}
\chapter*{Introduction}\label{sec:introduction}
\addcontentsline{toc}{chapter}{Introduction}
In this thesis we are interested in applying distributed estimation, control and optimization techniques to enable a group of quadcopters to fly through openings. The quadcopters are assumed to be equipped with a simulated bearing and distance sensor for localization. Some quadcopters are designated as leaders who carry global position sensors. We assume quadcopters can communicate information with each other. Under these assumptions, the goal of project was achieved by completing the following tasks:

\begin{enumerate}
\item Estimate global positions from distance and bearing measurements.
\item Form and maintain desired shapes.
\item Change the scale of the formation.
\item Plan collision free trajectories to pass through openings.
\end{enumerate}

The thesis is organized in the following way to address above challenges: Chapter~\ref{ch:graph} presented preliminary graph theory that is frequently referenced throughout the thesis. Chapter~\ref{ch:system} gave an overview of the system set-up and explains the principle of bearing and distance sensor. Chapter~\ref{ch:observer} introduced the distributed observer for swarm localization from the bearing and distance sensor measurements. Chapter~\ref{ch:control} discussed the distributed control for formation maintaining and chapter~\ref{ch:scale} proposed a formation scale estimation method for time varying formation scale. Finally Chapter~\ref{ch:optimization} proposed two scalable trajectory optimization algorithms to plan collision-free trajectories through the openings and gave an overview of the constraints.

\cleardoublepage
\chapter{Graph Theory}\label{ch:graph}
The concepts presented in this chapter are mainly summarized from the book~\cite{bullo2009distributed}. 
\section{Graphs}
\paragraph{Undirected graph}
An undirected graph is a tuple $G=(V,E)$, where $V$ is a node list and $E$ is a set of edges of unordered pairs of nodes. An unordered edge is a set of two nodes $\{i,j\}$, $\forall i,j \in V$ and $i\neq j$. If $\{i,j\}\in E$, then $i,j$ are called neighbors. Let $N_i$ denote the set of neighbors of $i$.

\paragraph{Directed graph}
A directed graph (diagraph) is a tuple $G=(V,E)$, where $V$ is a node list and $E$ is a set of edges of ordered pairs of nodes. An ordered pair $(i,j)$ denotes an edge from $i$ to $j$. $i$ is called \textit{in-neighbor} of $j$ and $j$ is called \textit{out-neighbor} of $i$. Let $N^{out}_i$ denote the set of \textit{out-neighbor} of $i$.

\paragraph{Path}
A path of undirected graph is an ordered sequence of nodes such that any pair of consecutive nodes is an edge of the graph. 

\paragraph{Directed path}
A directed path of a diagraph is an ordered sequence of nodes such that any pair of consecutive nodes is an edge of the diagraph.

\paragraph{Connectivity}
An undirected graph is \textit{connected} if between any pair of nodes there exist a path. A diagraph is \textit{strongly connected} if there exists a directed path between any pair of nodes.  

\paragraph{Weighted diagraph}
A weighted diagraph is a triplet $G=(V,E,\{a_e\}_{e\in E})$, where $(V,E)$ is a diagraph and $a_e$ is a strictly positive scalar of an edge $e\in E$. A weighted diagraph is undirected if for any edge $(i,j)\in E$, there exist an edge $(j,i)\in E$ and $a_{(i,j)}=a_{(j,i)}$. 

\section{Adjacency Matrix}
Given a weighted diagraph $G=(V,E,\{a_e\}_{e\in E})$, the adjacency matrix $A$ is defined as follows: 
\begin{equation}
a_{ij}=
\left\{\begin{array}{ll}
a_{(i,j)}, & \text{if $(i,j)\in E$}\\
0, & \text{otherwise}
\end{array}\right.
\end{equation}
The weighted out-degree matrix $D_{out}$ are defined by:
\begin{equation}
D_{out}=\text{diag}(A\onevect)
\end{equation}
\section{Laplacian Matrix}
The Laplacian matrix of the weighted diagraph $G=(V,E,\{a_{e}\}_{e\in E})$ is defined as:
\begin{align}
L&=D_{out}-A
\end{align}
and
\begin{align}
(Lx)_i&=\sum_{j\in N^{out}_{i}}a_{ij}(x_i-x_j)
\end{align}
If $G$ is undirected,
\begin{equation}
(Lx)_i=\sum_{j\in N_i}a_{ij}(x_i-x_j) \label{eq:Lx}
\end{equation}
and
\begin{align}
L\mathbf{1}=\mathbf{0}
\end{align}
For an undirected graph $G$, $L$ is symmetric. $\lambda_1=0$ is a simple eigenvalue of $L$ and all other eigenvalues of $L$ are real and positive:
\begin{align}
0=\lambda_1 \leq \lambda_2 \leq \cdots \leq \lambda_N
\end{align}
\section{Incidence Matrix}
Given an \textit{undirected weighted} graph $G=(V,E,\{a_e\}_{e\in E})$, number the edges of $G$ with a unique $e\in \{1,...,m\}$ and assign an arbitrary direction to each edge. The direction of an edge $(i,j)$ is from $i$ to $j$. $i$ is called the \textit{source} node of $(i,j)$ and $j$ is the \textit{sink} node. Then the incidence matrix B is defined element-wise as
\begin{equation}
B_{ie}=\left\{\begin{array}{ll}
+1,& \text{if node $i$ is the source node of edge $e$}\\ 
-1,& \text{if node $i$ is the sink node of edge $e$}\\
0,& \text{otherwise}
\end{array}
\right.
\end{equation}

The incidence matrix $B$ is used to compute the difference between the values of two nodes connected by an edge $e$ oriented from $i$ to $j$
\begin{equation}
(B^Tx)_e=x_i-x_j
\end{equation}
Thus $B^T\onevect=0$, where $\mathbf{1}=[1,...,1]^T$. Recall that 
\begin{equation}
(Lx)_i=\sum_{j\in N_i}a_{ij}(x_i-x_j)
\end{equation}
Then the Laplacian matrix $L$ is closely related to the incidence matrix $B$ through:
\begin{equation}
L=B\text{diag}(\{a_e\}_{e\in\{1,...,m\}})B^T
\end{equation}

The incidence matrix $B^T$ operates on $x$ to compute the difference between $x_i$ and $x_j$ of an edge $(i,j)$. The difference $(x_i-x_j)$ is then weighted by a diagonal element $a_{(ij)}$ of the diagonal matrix $\text{diag}(\{a_e\}_{e\in\{1,...,m\}})$. Finally $B$ picks and sums the weighted differences of all edges connecting the node $i$. The picking and summation properties of $B$ helps us to find an appropriate observer gain matrix in a later discuss about distributed observer design. Note that $L$ is symmetric because the weighted graph $G$ is undirected, i.e., $a_{(i,j)}=a_{(j,i)}$.
\cleardoublepage
\chapter{System Setup}\label{ch:system}
\section{Overview}
\begin{figure}[h]
\includegraphics[width=\textwidth]{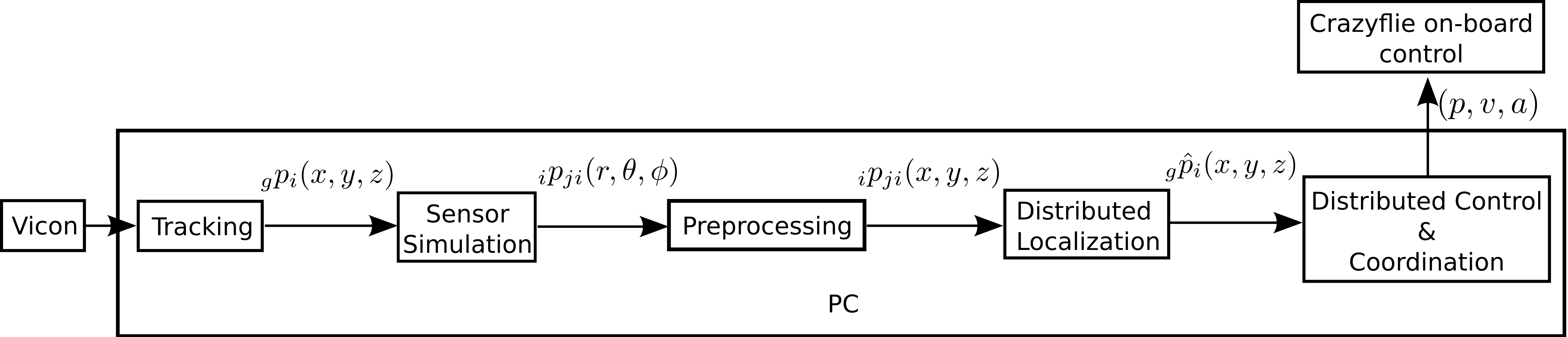}
\caption{Block diagram of the system}
\label{fig:block}
\end{figure}

An overview of the system is shown in Fig.~\ref{fig:block}. The testbed for this project is a nano quadcopter called Crazyflie 2.0. There are five main modules developed on PC in this project. A tracking module tracks the global position $_gp_i(x,y,z)$ of Crazyflie $i$. Then the tracked positions are used for simulating the local distance and bearing sensor measurement of Crazyflie $j$ measured by Crazyflie $i$. A preprocessing step follows to transform the simulated local measurement to global measurement from which the distributed localization module can estimate the global position of the Crazyflie. Based on the estimated positions, Crazyflies can apply distributed control to maintain a formation~\cite{oh2015survey} or plan trajectories to accomplish certain tasks~\cite{augugliaro2012generation}. In this chapter we discuss the hardware of the system and three modules: (a) Crazyflies tracking, (b) bearing and distance sensor simulation and (c) the preprocessing of the sensor measurements. 

\section{Hardware}
\begin{figure}
\centering
	\begin{subfigure}[h]{0.22\textwidth}
	\includegraphics[width=\textwidth]{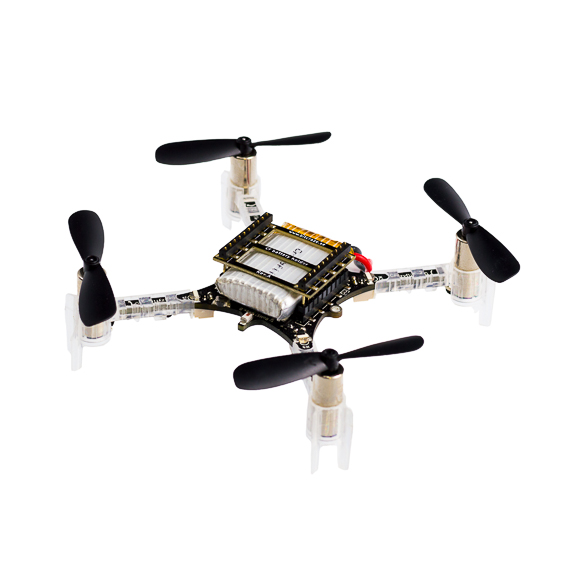}
	\caption{Crazyflie 2.0}
	\end{subfigure}
	\begin{subfigure}[h]{0.22\textwidth}
	\includegraphics[width=\textwidth]{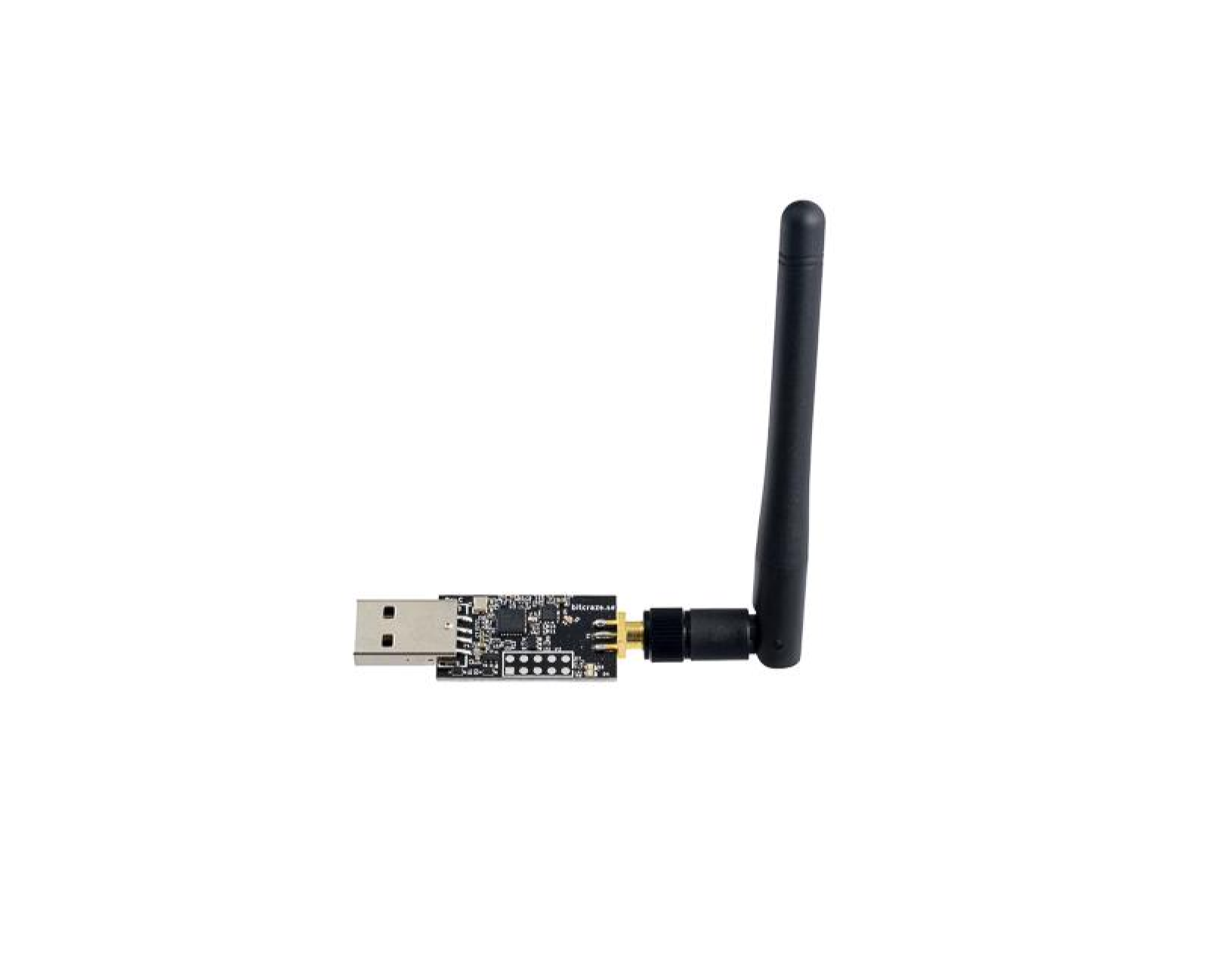}
	\caption{Crazyradio}
	\end{subfigure}
	\begin{subfigure}[h]{0.22\textwidth}
	\includegraphics[width=\textwidth]{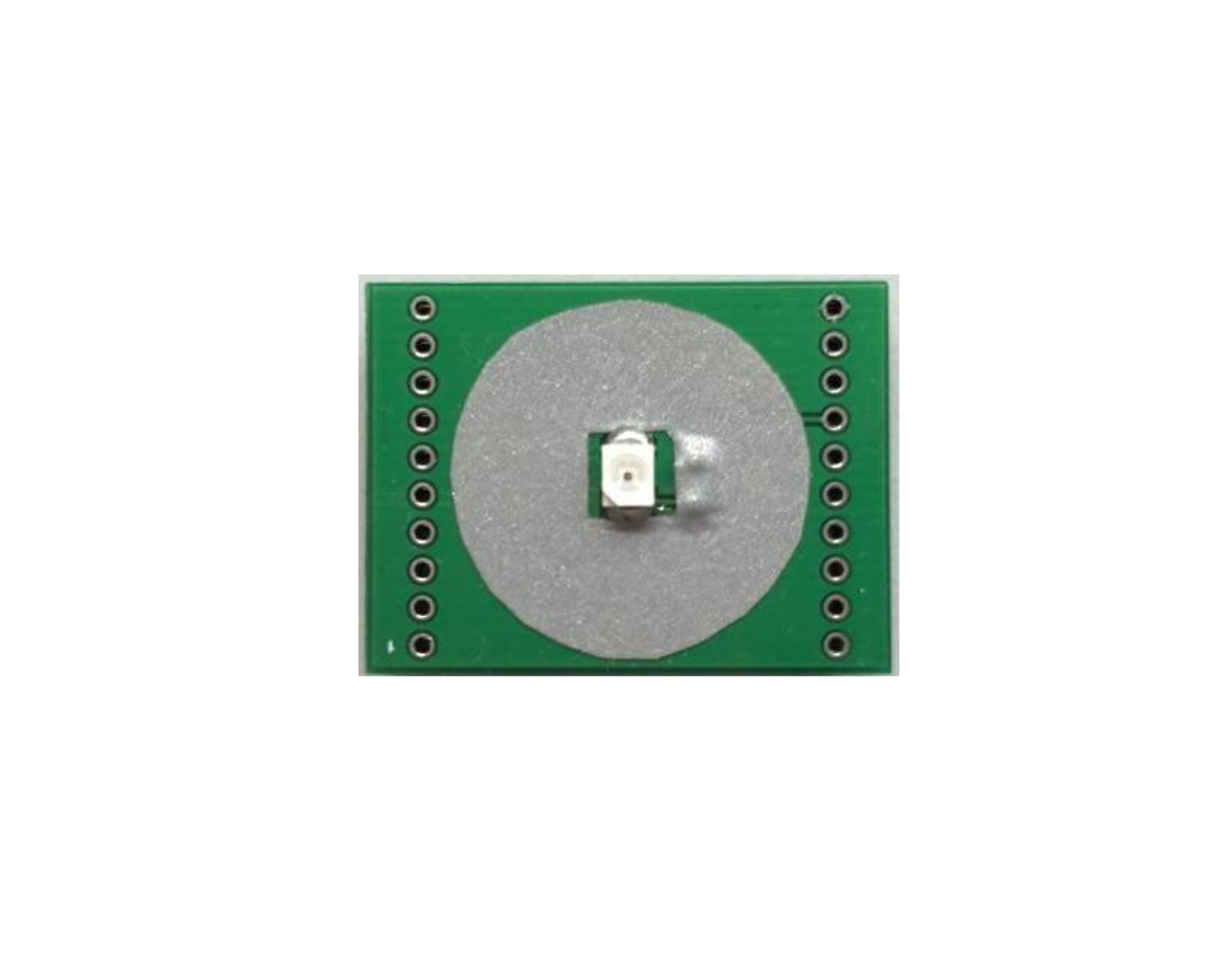}
	\caption{Infra-red led board}
	\end{subfigure}
	\begin{subfigure}[h]{0.22\textwidth}
	\includegraphics[width=\textwidth]{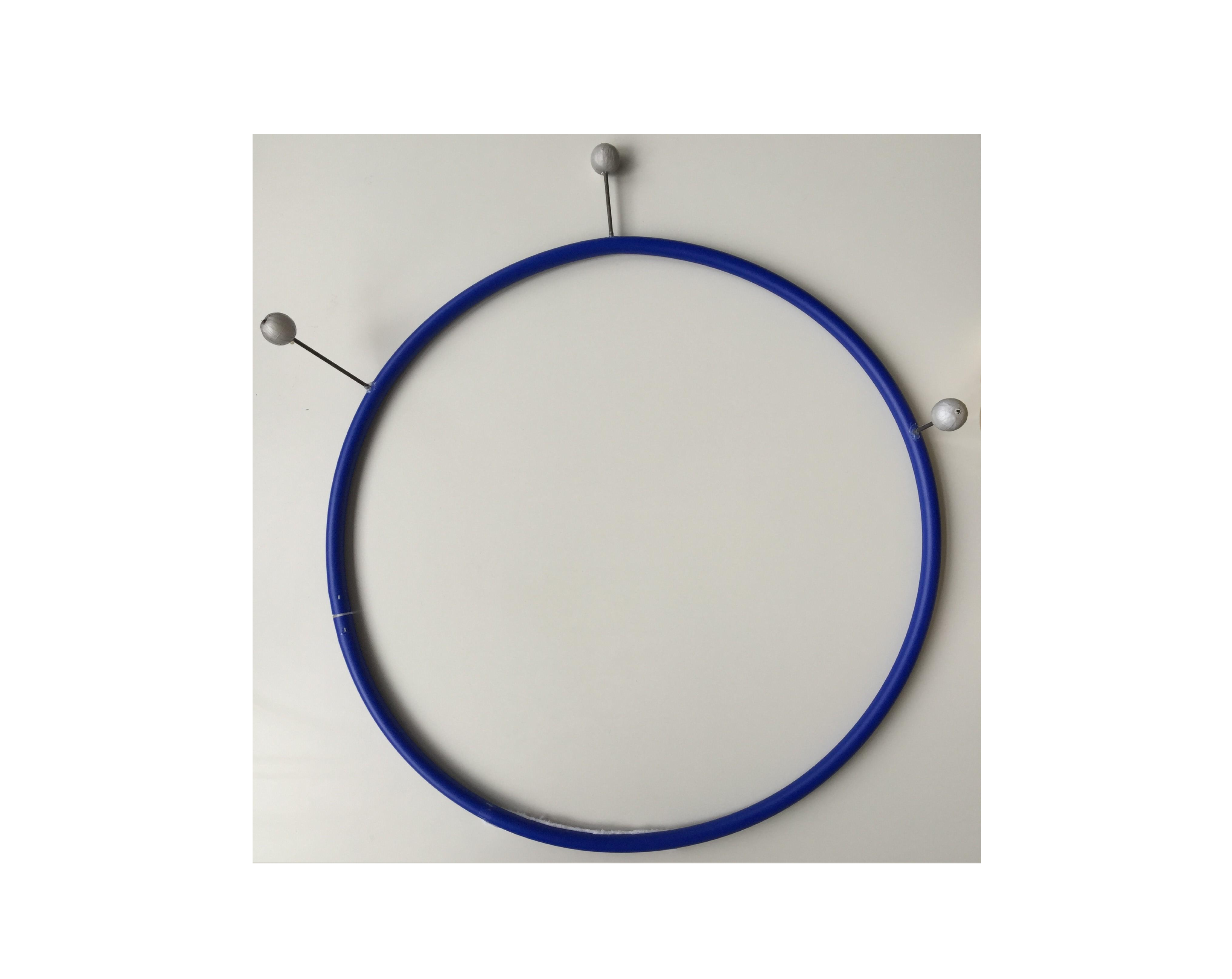}
	\caption{A ring with vicon markers attached}
	\end{subfigure}
	\caption{Available hardware}
	\label{fig:hardware}
\end{figure}

We have in total 4 Crazyflies and 1 Crazyradio for the communication between PC and Crazyflies. Since Crazyflies cannot directly communicate with each other, all communications were completed through PC. We used vicon system to track Crazyflies. On top of each Crazyflie, a infra-red led board is attached to strengthen the signal being observed by the vicon system. We used a hula loop to resemble a ring or opening. Three vicon markers were attached to the ring for tracking its position and orientation. The hardware used in this project are shown in Fig.~\ref{fig:hardware}.

\section{Crazyflies Tracking}
The vicon system outputs a set of unordered position measurements set $Z$ of which each element $z$ is the position measurement of a Crazyflie. Since the set is unordered, the correspondences of these measurements to the Crazyflies are unknown. In order to select the correct position measurement from the unordered measurements and be more robust to the missing measurements, we applied a kalman filter to track each Crazyflie. The Crazyflie is modelled as a random walk with unknown normally distributed acceleration $a_i$. Assume the state vector of the Crazyflie $i$ is:
\begin{align}
x_i=
\begin{bmatrix}
p_i \\
v_i
\end{bmatrix}
\end{align}
Then the random walk model of the Crazyflie is:
\begin{align}
p_i[k+1] &= p_i[k] + v_i[k]\delta t \\
v_i[k+1] &= v_i[k] + a_i[k]
\end{align}
where $\delta t=5$ms is the time interval between two consecutive measurements of vicon system. We applied the standard kalman filter to estimate the state $x$. Let $\hat{p}_{p,i}[k]$ and $\hat{v}_{p,i}[k]$ denote the prior updates of position and velocity of Crazyflie $i$ and $\hat{p}_{m,i}[k]$ and $\hat{v}_{m,i}[k]$ denote the measurement updates of Crazyflie $i$. Then the prediction step at time $k$ for Crazyflie $i$ is:
\begin{align}
\hat{p}_{p,i}[k+1] &= \hat{p}_{m,i}[k] + \hat{v}_{m,i}[k]\delta t \\
\hat{v}_{p,i}[k+1] &= \hat{v}_{m,i}[k] 
\end{align}

Since only the positions are measured, the measurement model is:
\begin{align}
z_i &= H_ix_i + w_i \\
&=\begin{bmatrix}
I & 0
\end{bmatrix}
\begin{bmatrix}
p_i\\
v_i
\end{bmatrix} + w_i\\
&=p_i + w_i
\end{align}

Before performing the measurement update step, the position measurement $z_i[k+1]$ of Crazyflie $i$ should be correctly picked from the unordered measurement set $Z$. It is selected by searching a measurement $z$ that has the smallest euclidean distance to the predicted position $\hat{z}_i[k+1]=\hat{p}_{p,i}[k+1]$:
\begin{align}
z_i[k+1] = \underset{z\in Z}{\text{argmin}}\|z-\hat{z}_i[k+1]\|^2_2
\end{align}

and the measurement update step at time $k$ is:
\begin{align}
\hat{p}_{m,i}[k+1] &= \hat{p}_{p,i}[k+1] + k_{p,i}(z_i[k+1]-\hat{z}_i[k+1])\\
\hat{v}_{m,i}[k+1] &= \hat{v}_{p,i}[k+1] + k_{v,i}(z_i[k+1]-\hat{z}_i[k+1]) 
\end{align}

where $k_{p,i}$ and $k_{v,i}$ are estimator gains. Thus we are able to track each Crazyflie from the unordered set of positions measurements. The tracked positions will be used for simulating the global position sensors and the bearing and distance sensor on Crazyflies.

\section{Bearing and Distance Sensor}
\subsection{Sensor Simulation}
\begin{figure}[h]
\centering
\includegraphics[width=0.4\textwidth]{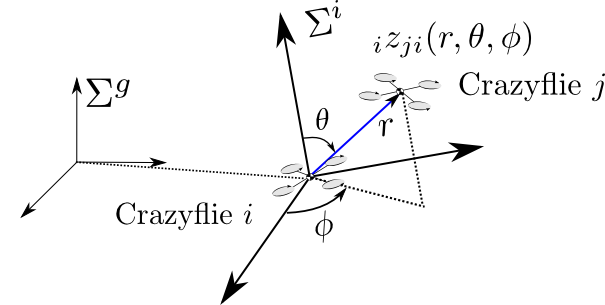}
\caption{Bearing and distance sensor}
\label{fig:sensor}
\end{figure}

As shown in Fig.~\ref{fig:sensor} the bearing and distance sensor attached on the Crazyflie $i$ measures the distance and angle $_iz_{ji}(r,\theta,\phi)$ of Crazyflie $j$ in its local coordinate $\Sigma^i$.

The local bearing and distance measurement $_iz_{ji}(r,\theta,\phi)$ is basically a Spherical coordinate representation. To simulate the bearing and distance sensor from vicon measurements, we first compute the global relative position measurement $_gz_{ji}$ from the tracked positions $_gz_i$ and $_gz_j$.
\begin{align}
_gz_{ji} = \text{${_gz_j}-{_gz_i}$}
\end{align}

and we can obtain the local representation of $_gz_{ji}$ in the local coordinate frame $\Sigma^i$ through the attitude matrix $_{ig}\hat{R}$ that is estimated by Crazyflie on-board attitude estimator:
\begin{align}
_iz_{ji} = {_{ig}\hat{R}^{-1}}_gz_{ji}
\end{align}

Let $_iz_{ji}=(x,y,z)$, then the local bearing and distance measurement $(r,\theta, \phi)$ can be found as:
\begin{align}
r &=\sqrt{x^2+y^2+z^2}\\
\theta &=\arccos \frac{z}{\sqrt{x^2+y^2+z^2}} \\
\phi &=\arctan \frac{y}{x}
\end{align}

\subsection{Preprocessing of Sensor Measurements}
\begin{figure}
\centering
	\begin{subfigure}[h]{0.2\textwidth}
		\includegraphics[width=\textwidth]{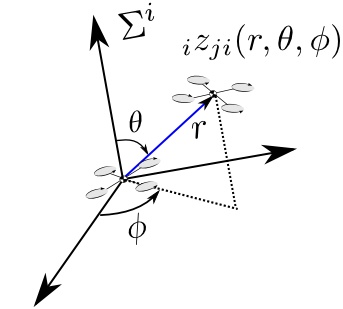}
		\caption{Local Spherical representation in $\Sigma^i$}	
	\end{subfigure}
	\begin{subfigure}[h]{0.18\textwidth}
		\includegraphics[width=\textwidth]{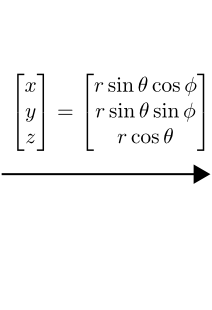}	
	\end{subfigure}
	\begin{subfigure}[h]{0.2\textwidth}
		\includegraphics[width=\textwidth]{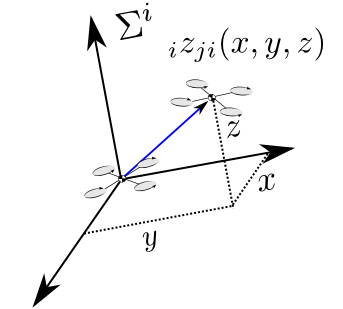}	
		\caption{Local Cartesian representation in $\Sigma^i$}
	\end{subfigure}
	\begin{subfigure}[h]{0.18\textwidth}
		\includegraphics[width=\textwidth]{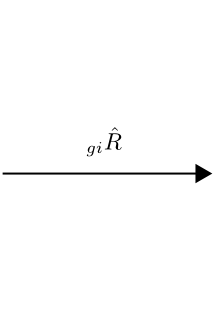}	
	\end{subfigure}	
	\begin{subfigure}[h]{0.2\textwidth}
		\includegraphics[width=\textwidth]{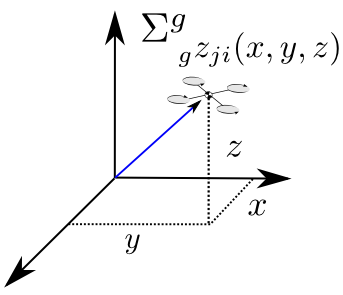}
		\caption{Global Cartesian representation in $\Sigma^g$}	
	\end{subfigure}
	\caption{Transformation of local bearing and distance measurement in Spherical representation to global Cartesian representation}	
	\label{fig:transformation}
\end{figure}

To estimate the global position $_gp_i$ from the local bearing and distance $_iz_{ji}(r,\theta,\phi)$ of neighbor $j$, a preprocessing step of transforming the Spherical representation to the global Cartesian representations is necessary. This transformation is illustrated in Fig.~\ref{fig:transformation}. For a distance and bearing measurement $_iz_{ji}(r,\theta,\phi)$ measured in Crazyflie $i$'s coordinate $\Sigma^i$, we first transform it to the local Cartesian representation $_iz_{ji}(x,y,z)$. 
\begin{align}
_iz_{ji}(x,y,z)
=
\prescript{}{i}{
\begin{bmatrix}
x\\
y\\
z
\end{bmatrix}}
=
\begin{bmatrix}
r\sin \theta \cos \phi\\
r\sin \theta \sin \phi\\
r\cos \theta
\end{bmatrix}
\end{align}

Furthermore, the local Cartesian representation can also be transformed to the global Cartesian representation through the attitude matrix $\prescript{}{gi}{\hat{R}}$. 
\begin{align}
_gz_{ji}(x,y,z)
=
\prescript{}{g}{
\begin{bmatrix}
x\\
y\\
z
\end{bmatrix}}
=
\prescript{}{gi}{\hat{R}}
\prescript{}{i}{
\begin{bmatrix}
x\\
y\\
z
\end{bmatrix}}
\end{align}

With the global representation of relative position $_gz_{ji}(x,y,z)$, the distributed observer is able to estimate the global positions of Crazyflies. Note that the attitude matrix $\prescript{}{gi}{\hat{R}}$ plays a crucial role here for the simulation and the preprocessing of the distance and bearing measurements.

\section{Parameters}
The parameters we used for tracking Crazyflie $i$, $i\in 1,...,N$ when the discretization time step $\delta t=5$ms are:
\begin{align}
&k_{p,i}=0.8\\
&k_{v,i}=0.0005
\end{align}
\cleardoublepage
\chapter{Distributed Observer}\label{ch:observer}
\section{Observer Design}
\begin{figure}
\centering
	\begin{subfigure}[h]{0.3\textwidth}
		\includegraphics[width=\textwidth]{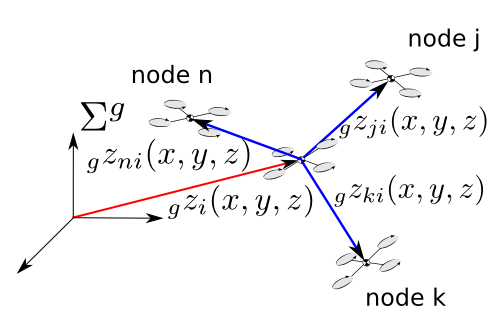}
		\caption{Global and relative position measurements}	
	\end{subfigure}
	~
	\begin{subfigure}[h]{0.3\textwidth}
		\includegraphics[width=\textwidth]{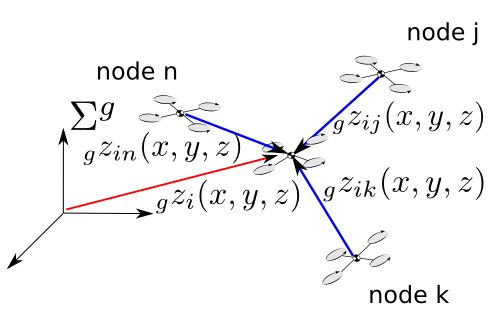}
		\caption{Reverse the direction of relative measurements}	
	\end{subfigure}
	~	
	\begin{subfigure}[h]{0.3\textwidth}
		\includegraphics[width=\textwidth]{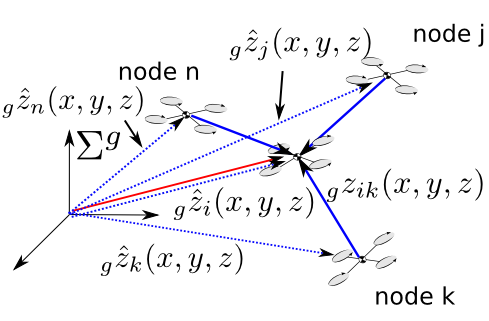}
		\caption{Fuse measurements with predictions}	
		\label{fig:fuse_measurement}
	\end{subfigure}
	\caption{Measurement update step}
	\label{fig:measurement_update}
\end{figure}

We assume that the underlying graph $G=(V,E)$ of the measurements is undirected and connected, that is, both Crazyflie $i$ and $j$ have the relative position measurement $_gz_{ij}$, after the preprocessing step. Let the edge $(i,j)$ of $G$ be numbered with a unique $e\in \{1,...,m\}$ and consider all $N$ Crazyflies are modeled as double-integrators in $n$-dimensional space (n=3):

\begin{equation}
\left\{\begin{array}{l}
\dot p_i = v_i, \\
\dot v_i = u_i,
\end{array}\right., i = \{1, \ldots, N\}
\end{equation}
The system dynamics in state space is:
\begin{equation}
\begin{bmatrix}
\dot p\\
\dot v
\end{bmatrix}
=	\underbrace{
    \begin{bmatrix}
    0 & I_{nN} \\
    0 & 0
    \end{bmatrix}}_\text{:=$A$}
    \begin{bmatrix}
    p \\
    v
    \end{bmatrix}
    +
    \underbrace{
    \begin{bmatrix}
    0 \\
    I_{nN}
    \end{bmatrix}}_\text{:=$B$}
    u
\end{equation}

As discussed in the previous chapter, the relative measurements are represented in global Cartesian coordinate $_gz_{ji}(x,y,z)$. From now on, the prescript $"g"$ and $(x,y,z)$ are dropped from $_gz_{ji}(x,y,z)$ for more concise notation. 

We are able to express the relative measurements $z_{ij}$ and the global measurements $z_{i}$ through the incidence matrix $B$  and the selection matrix $E$, where the selection matrix is defined as
\begin{align}
E&=\begin{bmatrix}
\cdots \underset{|}{\overset{|}{e_i}} \cdots
\end{bmatrix},\quad i\in V_g
\end{align}
and $E^Tz$ is a vector of all global measurements:
\begin{align}
(E^Tz)_k&=z_i,\quad i\in V_g
\end{align}
where $V_g$ is the node list of Crazyflies that carry global position sensors. $e_i$ is the $i$th column of the identity matrix $I_N$.  Let $x=[p^T,v^T]^T$ and $\hat{x}=[\hat{p}^T,\hat{v}^T]^T$. Then we decompose all measurements into relative and absolute measurements as follows:
\begin{align}
z &= Hx+w\\
&=
\begin{bmatrix}
B^T\kron I_n & 0\\
E^T\kron I_n	& 0
\end{bmatrix}
\begin{bmatrix}
p\\
v
\end{bmatrix}+w
\end{align}
The kronecker product $\kron$ generalizes system to n-dimentional space and $w$ is the vector of measurement noise. A typical state observer can be designed as:
\begin{align}
\dot{\hat{x}} &=A\hat{x}+Bu+L(z-\hat{z})\\
\hat{z} &= H\hat{x}\\
z &= Hx+w 
\end{align}

Usually a steady state optimal kalman filter $K_{\infty}$ will be used to design the gain $L$:
\begin{equation}
\dot{\hat{x}} =A\hat{x}+Bu+K_{\infty}(z-\hat{z})
\end{equation}

However the optimal kalman gain $K_\infty$ is a dense matrix that will fuse all measurements $z$ available in the network to update each Crazyflie $i$'s state $\hat{x}_i$~\cite{lu2015distributed}. This is not possible since the measurements and communications are only local and the local state observer of agent $i$ updates states only from the neighbors' relative measurements and from its own global position measurements:
\begin{align}
\begin{bmatrix}
\dot{\hat{p}}_i\\
\dot{\hat{v}}_i
\end{bmatrix}
&=
\begin{bmatrix}
0 & I_n\\
0 & 0
\end{bmatrix}
\begin{bmatrix}
\hat{p}_i\\
\hat{v}_i
\end{bmatrix}
+
\begin{bmatrix}
0\\
I_n
\end{bmatrix}
u_i
+
\left(\begin{bmatrix}
k_p & k_p\\
k_v & k_v
\end{bmatrix}
\kron I_n \right)
\begin{bmatrix}
\sum_{j\in N_i}k_{rp,ij}(z_{ij}-\hat{z}_{ij})\\
k_{gp,i}(z_i-\hat{z}_i)
\end{bmatrix}
\end{align}
where $k_{rp,i}>0$ and $k_{gp,i}\geq 0$ control the weights of the  measurements available to $i$ and $k_{gp,i}=0$ if Crazyflie $i$ does not have global position measurement. $k_p$ and $k_v$ were used to fine-tune the gains for updating positions and velocities. Note that $\sum_{j\in N_i}$ only sums local weighted relative measurements error $k_{rp,i}(z_{ij}-\hat{z}_{ij})$. As shown in Fig.~\ref{fig:fuse_measurement}, $(z_{ij}+\hat{z}_j)$ can be viewed as a global position measurement of $i$ and its difference with the prediction of global position $\hat{z}_i$ is the measurement error: 
\begin{align}
z_{ij}+\hat{z}_j-\hat{z}_i=z_{ij}-\hat{z}_{ij}
\end{align}
Let $D_{rp}=\text{diag}(\{k_{rp,e}\})_{e\in {1,...,m}}$, then $BD_{rp}B^T$ is a Laplacian matrix and
\begin{align}
(BD_{rp}B^T(z-\hat{z}))_i=\sum_{j\in N_i}k_{rp,ij}(z_{ij}-\hat{z}_{ij})
\end{align} 
Let $D_{gp}=\text{diag}(\{k_{gp,i}\})_{i\in V_g}$, then
\begin{align}
(ED_{gp}E^T(z-\hat{z}))_i=k_{gp,i}(z_i-\hat{z}_i)
\end{align}
Therefore the distributed observer can be written in a compact form as:
\begin{align}
\begin{bmatrix}
\dot{\hat{p}}\\
\dot{\hat{v}}
\end{bmatrix}
&=
\begin{bmatrix}
0 & I_{nN}\\
0 & 0
\end{bmatrix}
\begin{bmatrix}
\hat{p}\\
\hat{v}
\end{bmatrix}
+
\begin{bmatrix}
0\\
I_{nN}
\end{bmatrix}
u
+
L_1H
\begin{bmatrix}
z-\hat{z}
\end{bmatrix}
\end{align}
and the observer gain $L_1$ is:
\begin{align}
L_1&=
\begin{bmatrix}
k_p B D_{rp} & k_p E D_{gp}\\
k_v B D_{rp} & k_v E D_{gp}
\end{bmatrix}\kron I_n
\end{align}

\section{Stability}
To study the stability of above observer, let $e=[p^T,v^T]^T-[\hat{p}^T,\hat{v}^T]^T$. Then
\begin{equation}
\dot{e}
=
\underbrace{(A-L_1H)}_\text{:=$\mathcal{O}$}e
\end{equation}
and 
\begin{align}
\mathcal{O}=A-L_1H&=
\begin{bmatrix}
0 & I_{nN}\\
0 & 0
\end{bmatrix}
-
\begin{bmatrix}
k_p B D_{rp} & k_p E D_{ap}\\
k_v B D_{rp} & k_v E D_{ap}
\end{bmatrix}\kron I_n
\begin{bmatrix}
B^T & 0\\
E^T & 0
\end{bmatrix}\kron I_n\\
&=
\begin{bmatrix}
0 & I_{nN}\\
0 & 0
\end{bmatrix}-
\begin{bmatrix}
k_pBD_{rp}B^T+k_pED_{ap}E^T & 0\\
k_vBD_{rp}B^T+k_vED_{ap}E^T & 0
\end{bmatrix}\kron I_n\\
&=
\underbrace{\begin{bmatrix}
-k_pBD_{rp}B^T-k_pED_{ap}E^T & I_{N}\\
-k_vBD_{rp}B^T-k_vED_{ap}E^T & 0
\end{bmatrix}}_{:=\mathcal{O'}}\kron I_n
\end{align}

To ensure the observer estimates states properly, the error dynamics matrix $\mathcal{O}$ needs to be asymptotically stable. The kronecker product only add multiplicities of each eigenvalues and do not affect stability. We then find eigenvalues of $\mathcal{O'}$ to study the stability of $\mathcal{O}$ by solving $det(\lambda I_{2N}-\mathcal{O'})=0$.
\begin{align}
det(\lambda I_{2N}-\mathcal{O'})
&=
det(\begin{bmatrix}
\lambda I_{N} & 0\\
0 & \lambda I_{N}
\end{bmatrix}-
\begin{bmatrix}
-k_pBD_{rp}B^T-k_pED_{ap}E^T & I_{N}\\
-k_vBD_{rp}B^T-k_vED_{ap}E^T & 0
\end{bmatrix})\\
&=
det(\begin{bmatrix}
\lambda I_{N}+k_pBD_{rp}B^T+k_pED_{ap}E^T & -I_{N}\\
k_vBD_{rp}B^T+k_vED_{ap}E^T & \lambda I_{N}
\end{bmatrix})\\
&=
det(\lambda^2 I_{N}+\lambda k_pBD_{rp}B^T+\lambda k_pED_{ap}E^T + k_vBD_{rp}B^T+k_vED_{ap}E^T)\\
&=
det(\lambda^2 I_{N}+(\lambda k_p+k_v) (\underbrace{BD_{rp}B^T+ED_{ap}E^T}_\text{$:=\mathcal{T}$}))
\end{align}

Let $\mu_i$ be $i$th eigenvalue of $\mathcal{T}$, then
\begin{align}
det(\lambda^2 I_{N}+(\lambda k_p+k_v) \mathcal{T})=
\prod_{i}^{N}(\lambda^2+(\lambda k_p+k_v)\mu_i)=0
\end{align}
Then we can solve above equation to find $\lambda$
\begin{align}
\lambda^2+(\lambda k_p+k_v)\mu_i=0
\end{align}
The Routh-Hurwitz stability criterion tells us that if the coefficients of second order polynomial are all positive then the roots are in the left half plane. Therefore if $\mu_i>0$, then $\mathcal{O'}$ is Hurwitz. It is easy to verify that $\mathcal{T}$ is positive definite and thus $\mu_i>0$:
\begin{align}
x^T\mathcal{T}x&=x^T(BD_{rp}B^T+ED_{ap}E^T)x\\
&=\|\sqrt{D_{rp}} B^Tx\|^2_2+\|\sqrt{D_{ap}}E^Tx\|^2_2 \\
&\geq 0, \quad \forall x\in R^{3N}
\end{align}

Since $\|\sqrt{D_{rp}}B^Tx\|=0$ if and only if $x=\beta \mathbf{1}_{3N}=\beta(1,,,1)$, $\beta \in \Re$ and $\| E^T(\beta \mathbf{1}_{3N})\|>0$, 
\begin{align}
x^T\mathcal{T}x
&=\|\sqrt{D_{rp}} B^Tx\|^2_2+\|\sqrt{D_{ap}}E^Tx\|^2_2 > 0, \quad \forall x\in R^{3N}
\end{align}
Thus $\mu_i>0$ and $\mathcal{O'}$ is Hurwitz. The estimation error $\tilde{e}=[p^T,v^T]^T-[\hat{p}^T,\hat{v}^T]^T$ will asymptotically converge to zero. Thus all the agents are able to estimate positions and velocities, and we are ready to apply distributed control law to control the swarm.

\section{Parameters}
The observer gains to update position and velocity are:
\begin{align}
&k_p=0.8\\
&k_v=20.0
\end{align}
(a) If i measures global position, the observer gains used in this project are:
\begin{align}
k_{rp,ij}=k_{gp,i}=\frac{k_p}{N_i+1}, \quad \forall i\in V_g \\
k_{rv,ij}=k_{gv,i}=\frac{k_v}{N_i+1}, \quad \forall i\in V_g
\end{align}

(b) If i has no global position sensor $i\in V\setminus V_g$, then:
 \begin{align}
&k_{rp,ij}=\frac{k_p}{N_i}\\
&k_{gp,i}=0\\
&k_{rv,ij}=\frac{k_v}{N_i}\\
&k_{gv,i}=0
\end{align}

The parameters selected may not be optimal and need to be further tuned by trial and error in practice.
\cleardoublepage
\chapter{Distributed Control}\label{ch:control}
\section{Controller Design}
Let $p^*=[...,p^{*T}_i,...]^T$ and $v^*=[...,v^{*T}_i,...]^T$ be the desired global positions and velocities of all Crazyflies. Then the desired relative positions and velocities between $i$ and $j$ are $p^*_{ij}=p^*_i-p^*_j$ and $v^*_{ij}=v^*_i-v^*_j$. Among the Crazyflies, leaders are able to control the global positions $p^*_i$ and velocities $v^*_i$, whereas followers can only control the relative positions $p^*_{ij}$ and velocities $v^*_{ij}$. To maintain the formation, we propose implementing following formation control law on each Crazyflie $i$~\cite{oh2015survey}:
\begin{align}
u_i = & \quad u_{rp,i} + u_{rv,i} + u_{gp,i} + u_{gv,i}\\
u_i = &-\sum_{j\in N_i}k_{rp,ij}(p_i-p_j-p^*_{ij}) \\
	  &-\sum_{j\in N_i}k_{rv,ij}(v_i-v_j-v^*_{ij}) \\
      &-k_{gp,i}(p_i-p_i^*)	\\
      &-k_{gv,i}(v_i-v_i^*)
\end{align}
where $u_{rp,i}$ controls relative positions to achieve the desired formations, $u_{rv,i}$ controls relative velocities for the flocking behavior of the swarm, $u_{gp,i}$ controls the global position of the swarm, and $u_{gv,i}$ controls global velocity of the swarm. In addition, $k_{rp,ij}>0$, $k_{rv,ij}>0$ are the control gains for relative positions and velocities to neighbors. $k_{gp,i}\geq 0$, $k_{gv,i}\geq 0$ are control gains for the global positions and velocities. Only when Crazyflie $i$ is a leader, $k_{gp,i}> 0$, $k_{gv,i}> 0$. 

We assume the underlying graph to control positions and velocities be $G_{rp}=(V,E,\{k_{rp,e}\}_{e\in E})$ and $G_{rv}=(V,E,\{k_{rv,e}\}_{e\in E})$ respectively, which are both undirected and connected. Let $L_p$ and $L_v$ denote the Laplacian matrices for these two graphs. Let $G_p$ and $G_v$ be defined element-wise as: 
\begin{align}
(G_px)_i=k_{gp,i}x_i\\
(G_vx)_i=k_{gv,i}x_i 
\end{align}
and assume the ratio of control gains of position to velocity is constant $\alpha$, $\alpha \in \Re$: 
\begin{align}
k_{gp,i}&=\alpha k_{gv,i}\\
k_{rp,ij}&=\alpha k_{rv,ij}
\end{align}
Then the distributed control law in a compact form is:
\begin{align}
u&=K(x-x^*)\\
&=
    \begin{bmatrix}
    -(L_p\kron I_n)-(G_p\kron I_n) & -(L_v\kron I_n)-(G_v\kron I_n)
    \end{bmatrix}
    \begin{bmatrix}
    p-p^*\\
    v-v^*
    \end{bmatrix}\\
&=
    \begin{bmatrix}
    -(\alpha L_v\kron I_n)-(\alpha G_v\kron I_n) & -(L_v\kron I_n)-(G_v\kron I_n)
    \end{bmatrix}
    \begin{bmatrix}
    p-p^*\\
    v-v^*
    \end{bmatrix}
\end{align}
and we obtain the closed loop dynamics:
\begin{align}
\dot{x}&=Ax+Bu\\
\dot{x}&=
Ax+BK(x-x^*)\\
\dot{x}^*-\dot{x}&=\dot{x}^*-Ax-BK(x-x^*)\\
\dot{x}^*-\dot{x}&=\dot{x}^*-A(x-x^*+x^*)-BK(x-x^*)\\
\dot{x}^*-\dot{x}&=\dot{x}^*-Ax^*-A(x-x^*)-BK(x-x^*)\\
\dot{x}^*-\dot{x}&=Bu^*-(A+BK)(x-x^*)
\end{align}
Let $e=x^*-x$, $e_p=p^*-p$ and $e_v=v^*-v$, we then obtain the following error dynamics
\begin{align}
\dot{e}&=(A+BK)e+Bu^*\\
\begin{bmatrix}
  \dot e_p\\
  \dot e_v
  \end{bmatrix}
  &=
    \underbrace{\begin{bmatrix}
    0 & I_{nN} \\
    -(\alpha L_v\kron I_n)-(\alpha G_v\kron I_n) & -(L_v\kron I_n)-(G_v\kron I_n)
    \end{bmatrix}}_\text{$:=\mathcal{M}$}
      \begin{bmatrix}
      e_p \\
      e_v
      \end{bmatrix}
    +Bu^* \\
&=\left(\underbrace{\begin{bmatrix}
    0 & I_{N} \\
    -(\alpha L_v+\alpha G_v) & -(L_v+G_v)
\end{bmatrix}}_{:=\mathcal{M}'}\kron I_n \right)
\begin{bmatrix}
e_p\\
e_v
\end{bmatrix}
+Bu^*
\end{align}
One must make sure $\mathcal{M}'$ is Hurwitz such that $e_p$ and $e_v$ are bounded given $u^*$ is bounded. $u^*$ is the desired feed-forward input which is unknown.

\section{Stability}
Similar to the observer, we compute the eigenvalues of $\mathcal{M}'$ to test its stability:
\begin{align}
det(\lambda I_{2N} - \mathcal{M}')
&=
det(\begin{bmatrix}
	\lambda I_{N} & 0\\
	0 & \lambda I_{N}
\end{bmatrix}
-
\begin{bmatrix}
    0 & I_{N} \\
    -(\alpha L_v+\alpha G_v) & -(L_v+G_v)
\end{bmatrix})\\
&=
det(\begin{bmatrix}
\lambda I_N & -I_N\\
\alpha L_v+ \alpha G_v & \lambda I_N+L_v+G_v
\end{bmatrix}
)\\
&=
det(\lambda^2 I_N+\lambda L_v + \lambda G_v + \alpha L_v + \alpha G_v)\\
&=det(\lambda^2 I_N+(\lambda + \alpha)L_v + (\lambda + \alpha) G_v)\\
&=
det(\lambda^2 I_N+(\lambda + \alpha)(\underbrace{L_v+ G_v}_{:=\Gamma}))
\end{align}
Let $\gamma_i$ be the $i$th eigenvalue of $\Gamma$, then
\begin{align}
det(\lambda^2 I_N+(\lambda + \alpha)\Gamma)
&=\prod^N_i(\lambda^2 + (\lambda + \alpha)\gamma_i)=0
\end{align}
Again if $\gamma_i>0$ then $Re\{\lambda\}<0$.
\begin{align}
x^T(L_v+G_v)x &= x^T(L_v+G_v)x\\
&=x^TL_vx + x^TG_vx\\
&\geq 0
\end{align}
where $x^TL_vx\geq 0$ since $L_v$ is a Laplacian matrix which has non-negative eigenvalues and $L_vx=0$ if and only if $x=\beta \mathbf{1}_N$, $\beta\in \Re$. $x^TG_vx\geq 0$ because $G_v$ is a diagonal matrix with non-negative diagonal. Since $x^TG_vx>0$ when $x=\beta \mathbf{1}_N$, 
\begin{align}
x^T(L_v+G_v)x=x^TL_vx+x^TG_vx>0
\end{align}
Thus $\gamma_i>0$ and the eigenvalue $\lambda$ of $\mathcal{M}$ has negative real part. The error dynamics matrix $\mathcal{M}$ is asymptotically stable. 

\section{Parameters}
If $i$ is a leader $i\in V_g$, the control gains for position and velocity are:
\begin{align}
&k_{gp,i}=\frac{9.0}{N_i+1}\\
&k_{gv,i}=\frac{4.0}{N_i+1}\\
&k_{rp,ij}=\frac{9.0}{N_i+1}\\
&k_{rv,ij}=\frac{4.0}{N_i+1}
\end{align}
If $i$ is a follower $i\in V\setminus V_g$,
\begin{align}
&k_{gp,i}=0\\
&k_{gv,i}=0\\
&k_{rp,ij}=\frac{9.0}{N_i}\\
&k_{rv,ij}=\frac{4.0}{N_i}
\end{align}

Basically the above parameters mean that the relative position and velocity error to the neighbors are averaged to generate the final control output. This is simple and but may not be optimal. In practice they should be fine-tuned by trial and error.

\section{Discussion}
The distributed control in this project is only suitable for maintaining formations in free space. The collision avoidance was not considered in designing the control law. In literature, Lyapunov functions with infinite potential energy were often proposed to achieve collision avoidance~\cite{panagou2016distributed}. This is not realistic as physical systems have limited amount of actuation. This motivates us to apply optimization based trajectory generation to deal with complex environments and constraints as will be discussed in chapter~\ref{ch:optimization}. 
\cleardoublepage
\chapter{Formation Scale Estimation}\label{ch:scale}
\section{Estimator Design}
The distributed control maintains a constant shape of the formation by controlling the relative positions and velocities. Crazyflies need to know the scale factor of the relative positions and velocities to change the formation scale. In this chapter we discuss how to estimate the scale of the formation. 

Assume the center of the formation is $(p^c,v^c)$ and the desired relative position and velocities are $(p^r,v^r)$ with zero mean, i.e., $\sum_{e\in {1,...,m}}p^r_{e}=0$ and $\sum_{e\in {1,...,m}}v^r_{e}=0$. Then we are able to write the desired trajectories of the system as:
\begin{align}
\begin{bmatrix}
p^* \\
v^*
\end{bmatrix}
=
\begin{bmatrix}
p^c \\
v^c
\end{bmatrix}
+
\begin{bmatrix}
p^r \\
v^r
\end{bmatrix}
\end{align}
and
\begin{align}
v^r=\dot{p}^r
\end{align}
Sometimes the desired scale of the formation change over time, for example, to avoid collisions. Then the desired relative positions and velocities are not constant anymore and the desired trajectories can be rewritten as:
\begin{align}
\begin{bmatrix}
p^* \\
v^*
\end{bmatrix}
&=
\begin{bmatrix}
p^c \\
v^c
\end{bmatrix}
+
\begin{bmatrix}
p^r \\
\dot{p}^r
\end{bmatrix} \\
&=
\begin{bmatrix}
p^c \\
v^c
\end{bmatrix}
+
\begin{bmatrix}
s(t)\bar{p}^r \\
\dot{s}(t)\bar{p}^r
\end{bmatrix}
\end{align}
where $s(t)>0$ is the time varying scale and $\bar{p}^r$ is the relative positions when $s(t)=1$. Since we only allow leaders to have the information of the desired scale $s(t)$, the followers must communicate with neighbors to obtain this scale. One approach is that the scale $s(t)$ can be transmitted by leaders to their neighbors who in turn transmit $s(t)$ to their neighbors~\cite{coogan2011scaling}. However each follower needs to know among which of its neighbors there is a path to the leaders. This may not be robust in case the roles of leaders and followers may change and the communication may be lost. A better solution is again using distributed law to fuse all neighbors' information to estimate the scale regardless of the path to leaders~\cite{coogan2011scaling}. Assume the underlying communication graph $G$ is undirected and connected with weights $a_{ij}$ and $L_s$ is its Laplacian matrix. Let $V_g$ be the list of leaders who know the desired scale $s(t)$ and $s_{est,i}=[...,s_{est,i},...]^T$ be the scale estimate vector of all Crazyflies. The Crazyflie $i$ updates $s_{est,i}$ through:
\begin{align}
\dot{s}_{est,i} = -\sum_{j\in N_i}a_{ij}(s_{est,i}-s_{est,j})-g_i(s_{est,i}-s(t))
\end{align}
and
\begin{align}
\left\{\begin{aligned}
g_i&=g>0,\quad \text{if}\ i\in V_g\\
g_i&=0,\quad \text{otherwise}\end{aligned} \right. 
\end{align}
Let $G_s$ be a diagonal matrix and 
\begin{align}
(G_s)_i=g,\quad i\in V_g
\end{align}
Then we can write the estimation dynamics in compact form:
\begin{align}
\dot{s}_{est} = -L_ss_{est}-G_s(s_{est}-s\mathbf{1}_N)
\end{align}

\section{Stability}
Let $e:=s_{est}-s\mathbf{1}_N$. If $s(t)$ is varying slowly and $\dot{s}(t)$ is bounded the estimation error $e(t)$ is bounded from the bounded input bounded state theory:
\begin{align}
\dot{s}_{est} - \dot{s}\mathbf{1}_N
&= -L_ss_{est} -G_s(s_{est}-s\mathbf{1}_N) - \dot{s}\mathbf{1}_N\\
\dot{e} &= -(L_s+G_s)(s_{est}-s\mathbf{1}_N) - \dot{s}\mathbf{1}_N,\quad\textrm{since }L_s\mathbf{1}_N=0_N \\
\dot{e} &= -(L_s+G_s)e - \dot{s}\mathbf{1}_N
\end{align}
As before, $x^TL_sx\geq 0$ and $x^TL_sx=0$ if and only if $x=\beta \mathbf{1}_N$, $\beta\in \Re$ and $\mathbf{1}_N^TG_s\mathbf{1}_N>0$. Therefore $L_s+G_s$ is exponentially stable. 

Note that $\dot{p}^*=v^*$ and $\dot{p}^c=v^c$. Then
\begin{align}
p^* &= p^c + s(t)\bar{p}^r \\
v^* &= \dot{p}^c + \dot{s}(t)\bar{p}^r
\end{align}
Substitute $s(t)$ and $\dot{s}(t)$ with the estimated scale $s_{est}$ and $\dot{s}_{est}$, we obtain
\begin{align}
p^*_{est} &\approx p^c + (\textrm{diag}(s_{est})\kron I_n) \bar{p}^r \\
v^*_{est} &\approx \dot{p}^c + (\textrm{diag}(\dot{s}_{est})\kron I_n) \bar{p}^r
\end{align}
Then we are able to write the approximated desired trajectories as:
\begin{align}
\begin{bmatrix}
p^*_{est} \\
v^*_{est}
\end{bmatrix}
\approx
\begin{bmatrix}
p^c \\
v^c
\end{bmatrix}
+
\begin{bmatrix}
\textrm{diag}(s_{est})\kron I_n\\
\textrm{diag}(-(L_s+G_s)s_{est}+sG_s\textbf{1}_N)\kron I_n
\end{bmatrix}
\bar{p}^r 
\end{align}

Thus the approximated desired relative positions and velocities are obtained from the estimated scale factor $s_{est}$. Crazyflies then can apply the formation control law to maintain the estimated time varying desired relative positions and velocities. 

\section{Parameters}
We again average neighbors' estimates $s_{est,j}$, $j\in N_i$ to update $s_{est,i}$:

(a) If $i$ is a leader, $i\in V_g$, then
\begin{align}
a_{ij}=g_i=\frac{1}{N_i+1}
\end{align}

(b) If $i$ is a follower, $i\in V\setminus V_g$, then
\begin{align}
&a_{ij}=\frac{1}{N_i}\\
&g_{ij}=0
\end{align}
\cleardoublepage
\chapter{Distributed Trajectory Optimization}\label{ch:optimization}
\section{Centralized Trajectory Optimization}
Federico proposed a discrete time trajectory optimization method to compute collision-free trajectories in a centralized manner~\cite{augugliaro2012generation}. The optimization problem is:\\
\begin{align}
\begin{tabular}{c c}
{$\underset{x\in \Re^{3NK}}{\text{min}}$}&{$f_0=\|x\|^2_2$}\\
{\text{subject to}} & {$\begin{aligned}
A_{eq}x&=b_{eq}\\
A_{in}x&\preceq b_{in}
\end{aligned}$}
\end{tabular}
\end{align}

where $x\in \Re^{3NK}$ is the stacked acceleration vectors of all quadcopters. Let $T$ and $k$ denote the trajectory duration and discretization time step, then $K=\frac{T}{h}$.  The equality constraint $A_{eq}=b_{eq}$ represents initial and final positions and velocities of the quadcopters and the inequality constraint $A_{in}x\preceq b_{in}$ contains the convexified collision avoidance constraints and other physical constraints, e.g., actuator constraints. Since the jerks of quadcopter is related to the magnitude of body rates, we seek for a minimum jerk solution to reduce the aggressiveness of attitude changing during the flight~\cite{mueller2015computationally}:

\begin{align}\label{opt:full}
\begin{tabular}{c c}
{$\underset{x\in \Re^{3NK}}{\text{min}}$}&{$f_0=\|Dx\|^2_2$}\\
{\text{subject to}} & {$\begin{aligned}
A_{eq}x&=b_{eq}\\
A_{in}x&\preceq b_{in}
\end{aligned}$}
\end{tabular}
\end{align}

where $D$ is to compute forward difference of $x$ to approximate the derivative of acceleration $x$. The above centralized approach scales poorly with number of Crazyflies. There are $O(N^2)$ collision avoidance constraints and $O(N)$ optimization variables. In the following sections we discuss how to make the optimization more scalable.  

\section{Initial Solution}\label{sec:initialSolution}
To obtain the convexified collision avoidance constraints and the ring constraint $A_{in}x\preceq b_{in}$, we need either an initial guess or a previous solution. Here we discuss one  possible solution as the initial guess. 

\begin{figure}[h]
\centering
\includegraphics[width=0.5\textwidth]{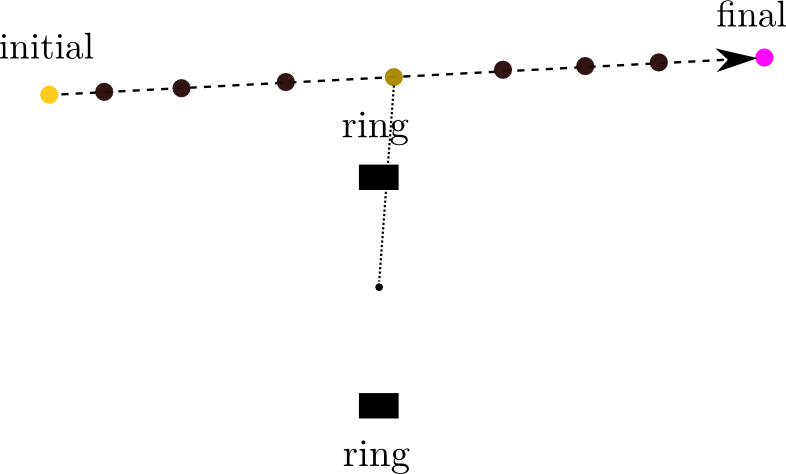}
\caption{Straight line solution}
\label{fig:straightLine}
\end{figure}

We first solve the optimization problem~\ref{opt:full} without $A_{in}\preceq b_{in}$ to find a straight line solution for each Crazyflie without considering the collision avoidance and ring constraints, which is shown in figure~\ref{fig:straightLine}

The straight line solution is used for finding a proper crossing time $k_{c}$ to impose the ring constraint. Then a velocity constraint at the position of the crossing time is imposed and the initial solution is refined such that it passes through the center of ring perpendicularly with a reasonable speed. 
\begin{figure}
\centering
\includegraphics[width=0.5\textwidth]{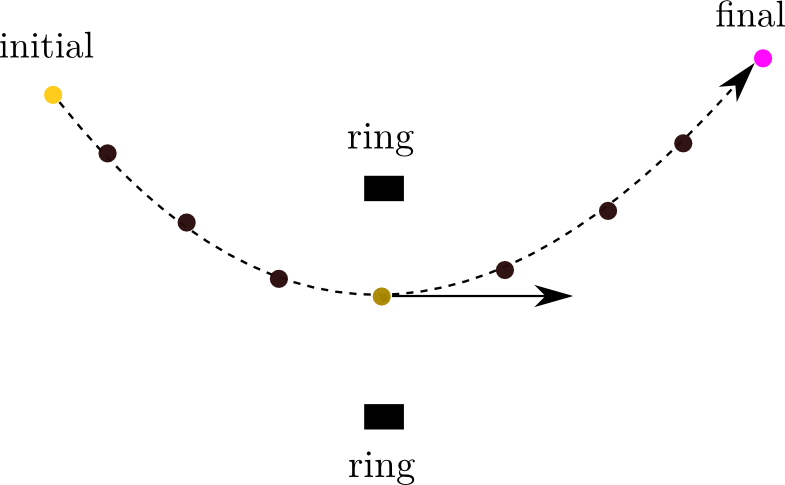}
\caption{Re-optimized solution passing through the center of ring}
\end{figure}

\begin{figure}
	\centering
	\includegraphics[width=0.1\textwidth]{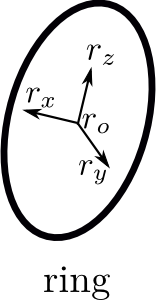}
	\caption{A coordinate $r_o-r_xr_yr_z$ is attached to the ring to represent the location and attitude of ring}
	\label{fig:ring}
\end{figure}

\section{Collision Avoidance Constraints}
There are two types of collisions considered to solve the optimization problem~\ref{opt:full}:

\subsection{Collisions between Crazyflie and ring} 
The ring resembles an opening that only its interior is allowed to pass through as illustrated in Fig.~\ref{fig:ring}. Therefore, Crazyflie should avoid hitting on or bypassing the ring. The ring object can be modelled as a convex circle constraint if we know the time to impose it when the Crazyflie crosses the ring. However it is unknown before solving the optimization problem. The following steps estimate a crossing time and convexify the ring constraint using an initial or previous $q$th solution $p^q_i[k]$:
\begin{enumerate}
	\item find $k^i_{c}=\underset{k\in 1,...,K}{\text{argmin}}\|p^q_i[k]-r_o\|^2_2$
	\item let $p^{q+1}_i[k^i_{c}]\in tube$
	\item let $p^{q+1}_i[k^i_{c}-1]\in leftCone$
	\item let $p^{q+1}_i[k^i_{c}+1]\in rightCone$
\end{enumerate}

We used a tube and two cones to approximate the ring constraint such that Crazyflie will avoid collisions before and after passing through it. The reason for imposing only two cone constraints is that it makes the constraints only be local near the ring and will not affect the optimization of trajectory far from the ring. This on one hand limits the number of constraints and on the other hand it also reduces the chances of infeasibility when the initial or final position that cannot be optimized are not within the cone, which results in an infeasible problem.

\paragraph{tube}
\begin{align}
	\left|{r}_y^T(p^{q+1}_i[k^i_{c}]-r_o)\right| \leq R_{tube} \\
	\label{eq:tube}
	\left|{r}_z^T(p^{q+1}_i[k^i_{c}]-r_o)\right| \leq R_{tube}
\end{align}
\paragraph{leftCone}
\begin{align}
	\left|{r}_y^T(p^{q+1}_i[k^i_{c}-1]-r_o)\right| \leq {r}_x^T(p^{q+1}_i[k^i_{c}-1]-r_o) \\
	\label{eq:leftCone}	
	\left|{r}_z^T(p^{q+1}_i[k_{c}-1]-r_o)\right| \leq {r}_x^T(p^{q+1}_i[k^i_{c}-1]-r_o)
\end{align}
\paragraph{rightCone}
\begin{align}
	\left|{r}_y^T(p^{q+1}_i[k^i_{c}+1]-r_o)\right| \leq - {r}_x^T(p^{q+1}_i[k^i_{c}+1]-r_o) \\
	\left|{r}_z^T(p^{q+1}_i[k^i_{c}+1]+r_o)\right| \leq - {r}_x^T(p^{q+1}_i[k^i_{c}+1]-r_o)
	\label{eq:rightCone}	
\end{align}

\begin{figure}[!h]
	\centering
	\begin{subfigure}[h]{0.5\textwidth}
		\includegraphics[width=\textwidth]{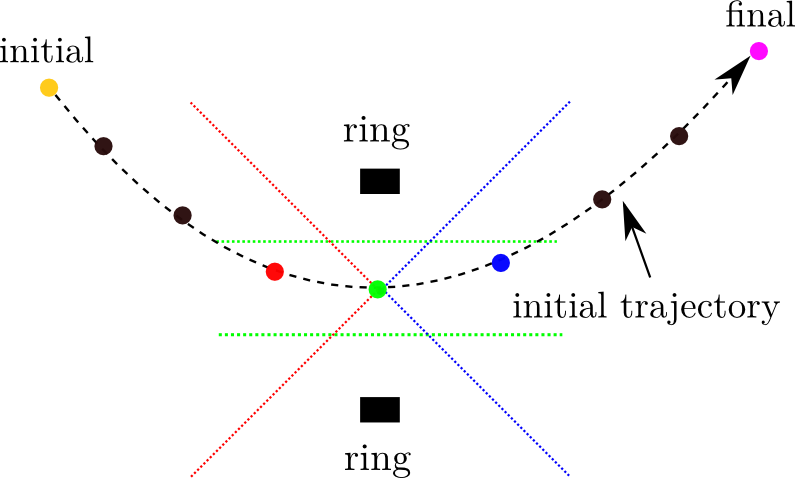}
		\caption{Impose tube (green), leftCone (red) and rightCone (blue) constraints to approximate the ring constraint}	
	\end{subfigure}
	~
	\begin{subfigure}[h]{0.5\textwidth}
		\includegraphics[width=\textwidth]{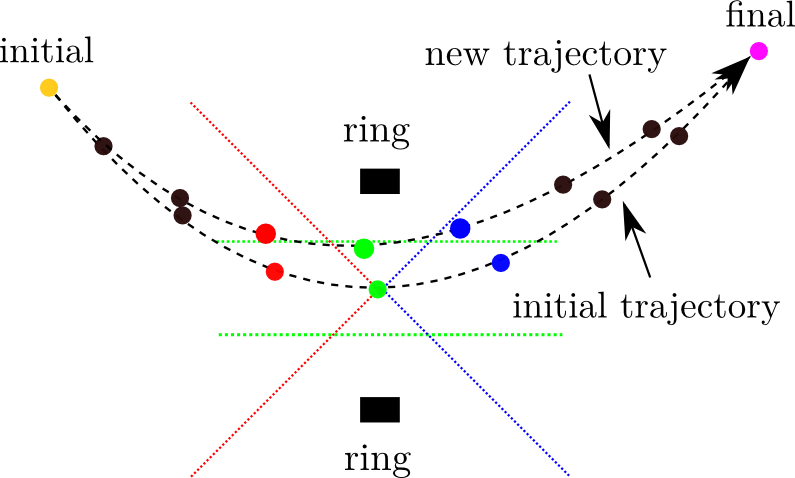}
		\caption{The resulting new trajectory after imposing the approximated ring constraint}	
	\end{subfigure}	
	\caption{Cross section diagram of ring constraint.}
\end{figure}

\subsection{Collisions between Crazyflies} 

To avoid collisions between Crazyflies themselves at each $k$, a safe distance margin $R_{collision}=0.3$m between Crazyflies' centers is enforced. This margin is large enough to tolerate certain control error when Crazyflies are tracking the trajectories. The collision avoidance constraint between Crazyflies $i$ and $j$ at time $k$ is a non-convex constraint:
\begin{align}
\|p_i[k]-p_j[k]\|_2 \geq R_{collision} \quad \forall i,j \in 1,...,N, \ i\neq j \label{eq:nonconvex_collision}
\end{align}

To convexify this constraint, again we need an initial guess or previous solution $p^q_i[k], p^q_j[k]$. Assume the new solutions are $p^{q+1}_i$ and $p^{q+1}_j$, then the convexified constraint is~\cite{augugliaro2012generation}:
\begin{align}
\eta^T(p^{q+1}_i[k]-p^{q+1}_j[k]) \geq R_{collision}, \quad \eta=\frac{p^{q}_i[k]-p^{q}_j[k]}{\|p^{q}_i[k]-p^{q}_j[k]\|_2}
\label{eq:collisionDouble}
\end{align}

\begin{figure}[h]
	\centering
	\begin{subfigure}[h]{0.3\textwidth}
		\includegraphics[width=\textwidth]{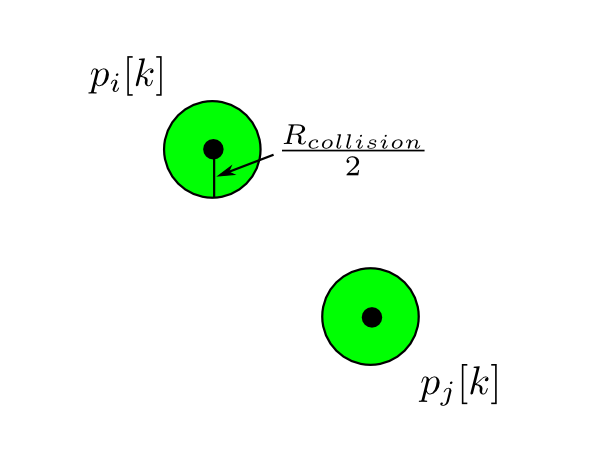}
		\caption{}
		\label{fig:collision}
	\end{subfigure}
	~
	\begin{subfigure}[h]{0.3\textwidth}
		\includegraphics[width=\textwidth]{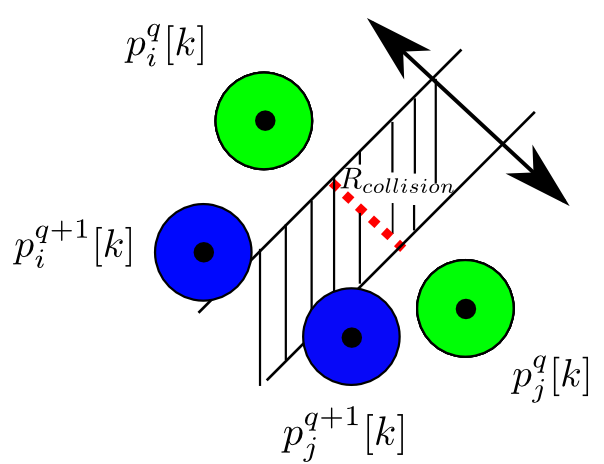}
		\caption{}
	\end{subfigure}
	~
	\begin{subfigure}[h]{0.3\textwidth}
		\includegraphics[width=\textwidth]{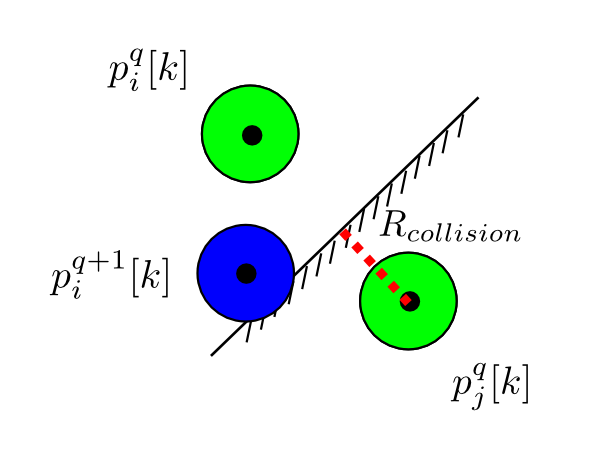}
		\caption{}
	\end{subfigure}
	\caption{Collision avoidance constraints. Fig.(a) illustrates the original non-convex collision avoidance constraint Eqn.~\ref{eq:nonconvex_collision}. Fig.(b) illustrates the convexified constraint Eqn.~\ref{eq:collisionDouble}. Fig.(c) illustrates the convexified constraint Eqn.~\ref{eq:collisionSinglei}.  }
	\label{fig:collision_comparison}
\end{figure}

Note that this convexified constraint assumes the optimization variables are $p^{q+1}_i[k],p^{q+1}_j[k]$. We can also optimize only $p^{q+1}_i[k]$ for Crazyflie $i$ and $p^{q+1}_j[k]$ for Crazyflie $j$ independently~\cite{chen2015decoupled}. 
\begin{align}
\eta^T(p^{q+1}_i[k]-p^{q}_j[k]) \geq R_{collision}, \quad \eta=\frac{p^{q}_i[k]-p^{q}_j[k]}{\|p^{q}_i[k]-p^{q}_j[k]\|_2}
\label{eq:collisionSinglei}
\\
\eta^T(p^{q}_i[k]-p^{q+1}_j[k]) \geq R_{collision}, \quad \eta=\frac{p^{q}_i[k]-p^{q}_j[k]}{\|p^{q}_i[k]-p^{q}_j[k]\|_2}
\label{eq:collisionSinglej}
\end{align}
As shown in Fig.~\ref{fig:collision_comparison}, the convexified constraints Eqn.~\ref{eq:collisionDouble} and Eqn.~\ref{eq:collisionSinglei} are different. Eqn.~\ref{eq:collisionDouble} is a relative constraint that the infeasible region can move along the direction of $\eta$, whereas Eqn.~\ref{eq:collisionDouble} is an absolute constraint that the infeasible region is static. Clearly the collision constraints convexified with Eqn.~\ref{eq:collisionDouble} have larger feasible regions.

\section{Distributed Constrained Convex Optimization}
After discussing constructing initial solution and convexfication of collision constraints, we begin discussing the method to make the optimization problem more scalable. Recently there is an increasing interest in enabling agents to cooperatively solve the following distributed constrained convex optimization:
\begin{align}\label{opt:dis}
\begin{tabular}{c c}
{$\underset{x}{\text{min}}$}&{$f_0=\overset{N}{\underset{i=1}{\sum}} f_i(x)$}\\
{\text{subject to}} & {$\begin{aligned}
x\in X\overset{\triangle}{=}\overset{N}{\underset{i=1}{\cap}}X_i
\end{aligned}$}
\end{tabular}
\end{align}
where each $f_i$ is a local objective function and $X_i$ is a local feasible set. The important observation here is that both the objective function $f_0$ and constraint set $X$ are decomposed as individual local objective functions and feasible sets. The distributed projected subgradient algorithm to solve the above problem is~\cite{lin2016distributed}:
\begin{align}\label{alg:dis_proj_sub_alg}
x_i(m+1) = P_{X_i}\left[\sum_{j\in N_i\cup i}a_{ij}x_j(m) - \alpha_m d_i(m)\right]
\end{align}
where $x_i(m)\in \Re^{3NK}$ is the local estimate of $x$ at time m, $a_{ij}$ is an entry of adjacency matrix $A$ of the underlying communication graph $G$, $\alpha_m>0$ is the step size of subgradient algorithm at time $m$, and $d_i(m)$ is a subgradient of local objective function at $\sum_{j\in N_i\cup i}a_{ij}x_j(m)$. $P_{X_i}[x]$ is the projection of $x$ onto $X_i$, i.e. $P_{X_i}(x)=\textrm{argmin}_{\bar x\in X_i}\|\bar x-x\|$. The distributed projected subgradient algorithm converges as $m\rightarrow \infty$:
\begin{align}
\lim_{m\rightarrow \infty}\|x_i(m)-x_j(m)\|^2_2=0
\end{align}
under the following assumptions~\cite{lin2016distributed}:
\begin{enumerate}
\item $X$ is nonempty and has nonempty interior.
\item The graph $G$ is fixed and strongly connected.
\item $a_{ij}>\eta$, $0<\eta<1$.
\item $A$ is doubly stochastic.
\item $X$ is compact.
\item $\sum^{+\infty}_{m=0}\alpha_m=+\infty$ and $\sum^{+\infty}_{m=0}\alpha^2_m<\infty$
\end{enumerate}

\begin{remark}
The $x_i(m)$ is a local estimate of $x$ at time m. The termination condition of the algorithm is that all the local estimates converge sufficiently close to each other, i.e., $\|x_i(m)-x_j(m)\|^2_2<\epsilon$, and we may assume at most $M$ iterations the algorithm can converge.
\end{remark}
\begin{remark} There are three operations involved in this algorithm. (a) $\sum_{j\in N_i\cup i}a_{ij}x_j(m)$ is a standard distributed averaging step such that $x_i(m)$ and $x_j(m)$ reach consensus as $m\rightarrow \infty$. (b) $-\alpha_m d_i(m)$ is a subgradient step to reduce the local objective function value. (c) $P_{X_i}[x]$ projects the averaged and subgradient subtracted solution to the local feasible set $X_i$. The combined three steps allows the local solution $x_i(m)$ to reach consensus with neighbors asymptotically as $m \rightarrow \infty$ and cooperatively reduce the value of objective function $f_0=\sum^N_{i=1} f_i(x)$ while satisfying their own local constraint $X_i$ after the projection $P_{X_i}[x]$. 
\end{remark}

In order to solve Problem~\ref{opt:full}, we express it in the form of problem~
\ref{opt:dis} as:
\begin{align}\label{opt:dis_jerk_collision}
\begin{tabular}{c l}
{$\underset{x\in \Re^{3NK}}{\text{min}}$}&{$f_0=\overset{N}{\underset{i=1}{\sum}} \|D_ix\|^2_2$}\\
{\text{subject to}} & {$\begin{aligned} 
x\in X \overset{\triangle}{=}\overset{N}{\underset{i=1}{\cap}} X_i, \quad X_i=\left\{ \begin{aligned}
&A_{eq}\ x=b_{eq} \\
&A_{in,i}\ x\preceq b_{in,i}
\end{aligned}\right.
\end{aligned}$}
\end{tabular}
\end{align}

where $D_i$ computes the jerk of Crazyflie $i$, $A_{eq}\ x=b_{eq}$ is the initial to final state condition of \textit{all} Crazyflies, and $A_{in,i}\ x\preceq b_{in,i}$ includes the convexified collision avoidance constraints of Crazyflie $i$ with the remaining $N-1$ Crazyflies and other constraints local to $i$, e.g., $a_{min,i}\preceq x_i \preceq a_{max,i}$.

\begin{remark} The assumption 1 that $X$ has nonempty interior does not strictly hold for problem~\ref{opt:dis_jerk_collision} as the equality constraint $A_{eq}\ x=b_{eq}$ is present. However experiments show that the algorithm still converges. One reason may be that all the Crazyflies share the same equality constraint $A_{eq}\ x= b_{eq}$, which does not affect the convergence of algorithm that assumes only inequality constraints exist. The assumption 3 and 4 can be guaranteed by setting $a_{ij}=\frac{1}{N_i+1}$, which also simplifies the algorithm design. The assumption 5 is satisfied because of the actuator constraint $a_{min}\preceq x \preceq a_{max}$ is compact. Finally in assumption 6, $\alpha_m$ is the step size of subgradient used for decreasing the objective function value. Experiments show that the algorithm converges faster when $\alpha_m$ is set close to 0. This is because the algorithm is solely trying to find a feasible point of $X$ without too much perturbation from the operation of subtracting subgradients. Since we are more interested in quickly finding a feasible solution rather than its optimality, we set $\alpha_m=0$ for fastest consensus rate. Although setting $\alpha_m=0$ violates assumption 6, experiments demonstrate that the algorithm always converges. 
\end{remark}

\begin{figure}
\centering
\includegraphics[width=0.6\textwidth]{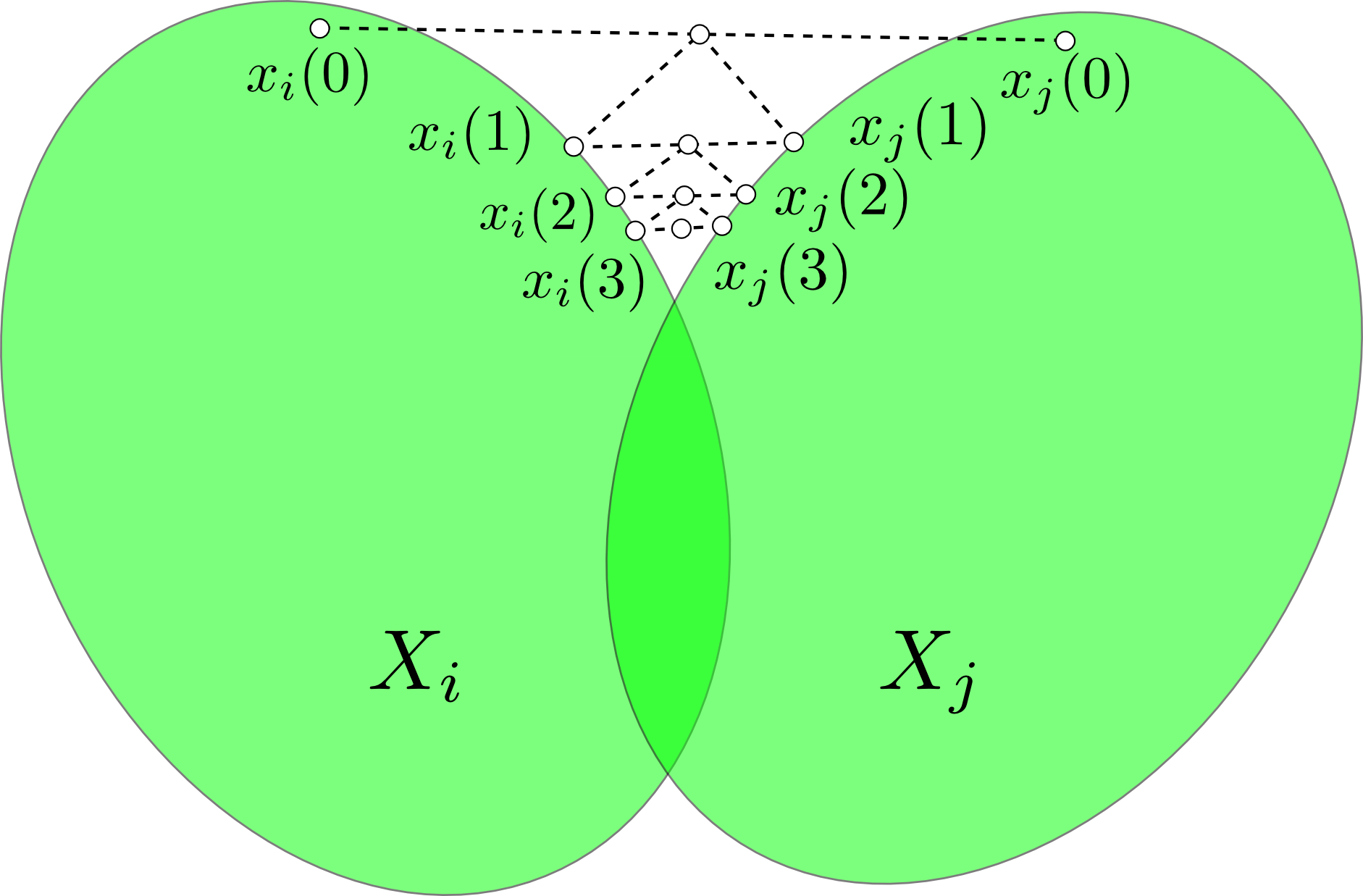}
\caption{Distributed projected subgradient algorithm for two agents with $\alpha_m$ set to 0. The algorithm begins with two initial solutions $x_i(0)$ and $x_j(0)$ in their own feasible sets $X_i$ and $X_j$ and are progressively averaged and projected to reach consensus in the intersection of $X_i$ and $X_j$.}
\label{fig:alternativeProjection}
\end{figure}

Assuming $a_{ij}=\frac{1}{N_i+1}$ and $\alpha_m=0$, the distributed projected subgradient algorithm~\ref{alg:dis_proj_sub_alg} becomes:
\begin{align}\label{alg:dis_proj_alg}
x_i(m+1) = P_{X_i}\left[\sum_{j\in N_i\cup i}\frac{1}{N_i+1}x_j(m)\right]
\end{align}
Fig.~\ref{fig:alternativeProjection} illustrates the process of two agents running the algorithm Eqn.~\ref{alg:dis_proj_alg}. The solutions $x_i(m)$ and $x_j(m)$ asymptotically converge to the intersection $X_i\cup X_j$ and reach consensus as $m \rightarrow \infty$.

\begin{remark}
The significance of distributed projected algorithm Eqn.~\ref{alg:dis_proj_alg} is that (a) the constraint set $X$ is decomposed to $X_i$ and distributed to each Crazyflie. Thus the number of constraints for each Crazyflie is small. (b) the algorithm runs in parallel. Compared to the centralized problem~\ref{opt:full} where the number of constraints of $X$ is of order $O(N^2)$, the number of constraints of $X_i$ of the problem~\ref{alg:dis_proj_alg} is equal to $N-1$ (because there are $N-1$ other Crazyflies for collision avoidance). The optimization may need to be solved $M$ times until convergence. As the solving time for convex problem increases quadratically with the number of inequalities, the runtime of problem~\ref{opt:full} and algorithm~\ref{alg:dis_proj_alg} are of order $O(N^4)$ and $O(M^2N^2)$.
\end{remark}

\begin{figure}
\centering
\includegraphics[scale=0.5]{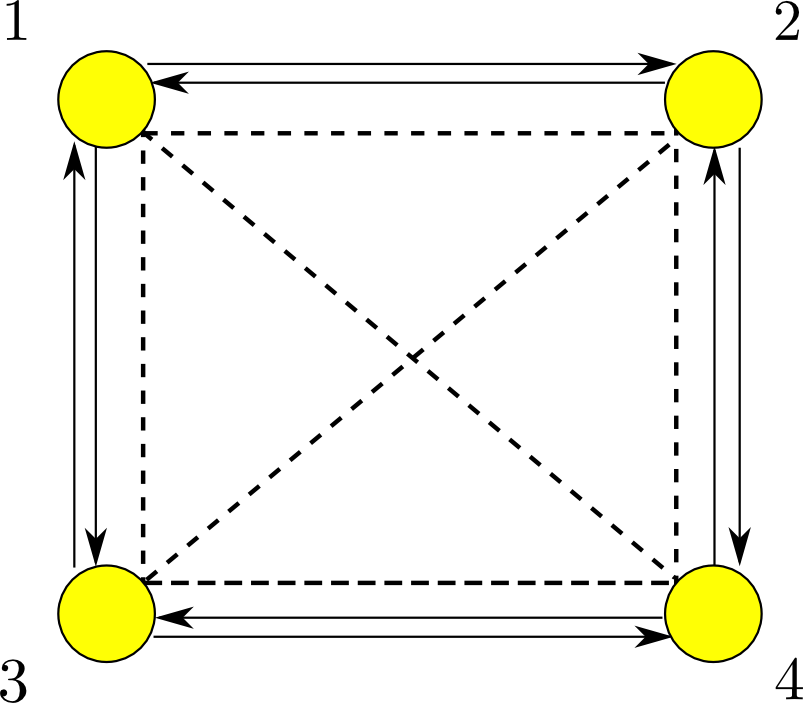}
\caption{An example of underlying communication graph for 4 Crazyflies. The arrows indicate the direction of message passing. The dashed lines indicate the existence of collision avoidance constraint between two Crazyflies.}
\label{fig:4AgentsGraph}
\end{figure}

\begin{remark}
Although $X_i$ of Crazyflie $i$ contains convexified collision avoidance constraints with the remaining $N-1$ Crazyflies, it does not mean it has to have established communication channels with the rest $N-1$ Crazyflies. Fig.~\ref{fig:4AgentsGraph} illustrates an example that 4 Crazyflies are running the distributed algorithm in parallel. There are 3 convexified collision constraints in any Crazyflie $i$'s local feasible set $X_i$. However each Crazyflie is communicating with only 2 Crazyflies. For example, Crazyflie 1 will only receive $x_2(m)$ and $x_3(m)$ at each time m. 
\end{remark}
\begin{remark}
The collision avoidance constraint for any pair of Crazyflies is included in both Crazyflies' constraint sets. For example in Fig.~\ref{fig:4AgentsGraph}, the constraint sets of both Crazyflie 1 and 4 include the collision constraint pair (1,4). This is redundant for the algorithm to converge. It is sufficient to let it be included in one of feasible sets. Nonetheless, this requires a protocol for constraints assignment. For simplicity, the collision constraint for a pair of Crazyflies is included in both feasible sets. 
\end{remark}

\begin{remark}
In order to convexifiy the collision avoidance constraint, before running distributed optimization, the initial solutions must be shared in the network. For the example in Fig.~\ref{fig:4AgentsGraph}, Crazyflie 1 needs to know the initial solution of Crazyflie 4 to convexify the collision avoidance constraint between 1 and 4. This is accomplished through the communication path 4-2-1 or 4-3-1. This may cause significant delay when the network is large.
\end{remark}

\section{Distributed Trajectory Optimization}\label{sec:Dis_Tra_Opt}
\subsection{Observations}
The distributed projected algorithm~\ref{alg:dis_proj_alg} successfully reduces the problem size from $O(N^2)$ to $O(N)$. However, even the problem size is $O(N)$, it will be quickly become not scalable as $N$ grows because:
\begin{enumerate}
\item To linearize collision avoidance constraint, the initial solutions of all Crazyflies must be shared in the network, which may be time-consuming when the network is large.
\item The dimension of $x_i$ is equal to $3NK$. Therefore, the communication time of $x_i$ after each optimization may approximately linearly grow with $N$.
\item The number of collision avoidance constraints of each Crazyflie is equal to $N-1$. As $N$ becomes large, the optimization will inevitably be intractable.
\end{enumerate}

\begin{figure}
\centering
	\begin{subfigure}[h]{0.42\textwidth}
	\includegraphics[width=\textwidth]{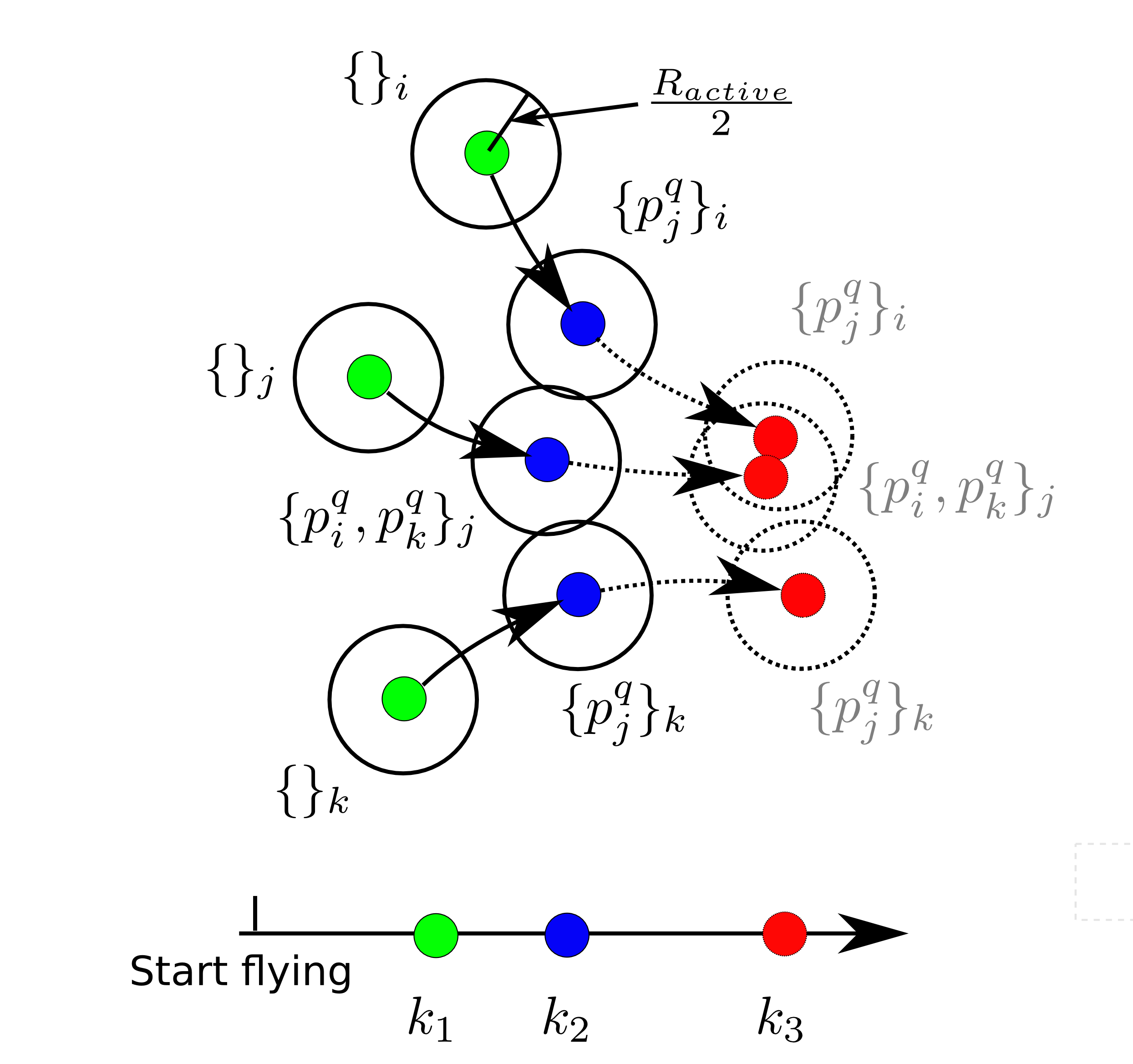}
	\caption{At $k_2$, $i,j$ detect collision at $k_3$}
	\label{fig:beforeOpt}
	\end{subfigure}
	~
	\begin{subfigure}[h]{0.1\textwidth}
	\includegraphics[width=\textwidth]{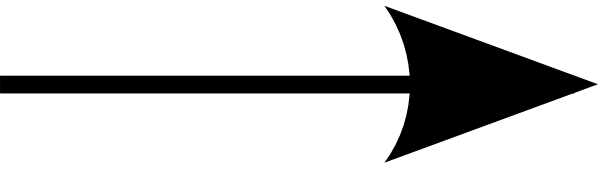}
	\end{subfigure}
	~
	\begin{subfigure}[h]{0.42\textwidth}
	\includegraphics[width=\textwidth]{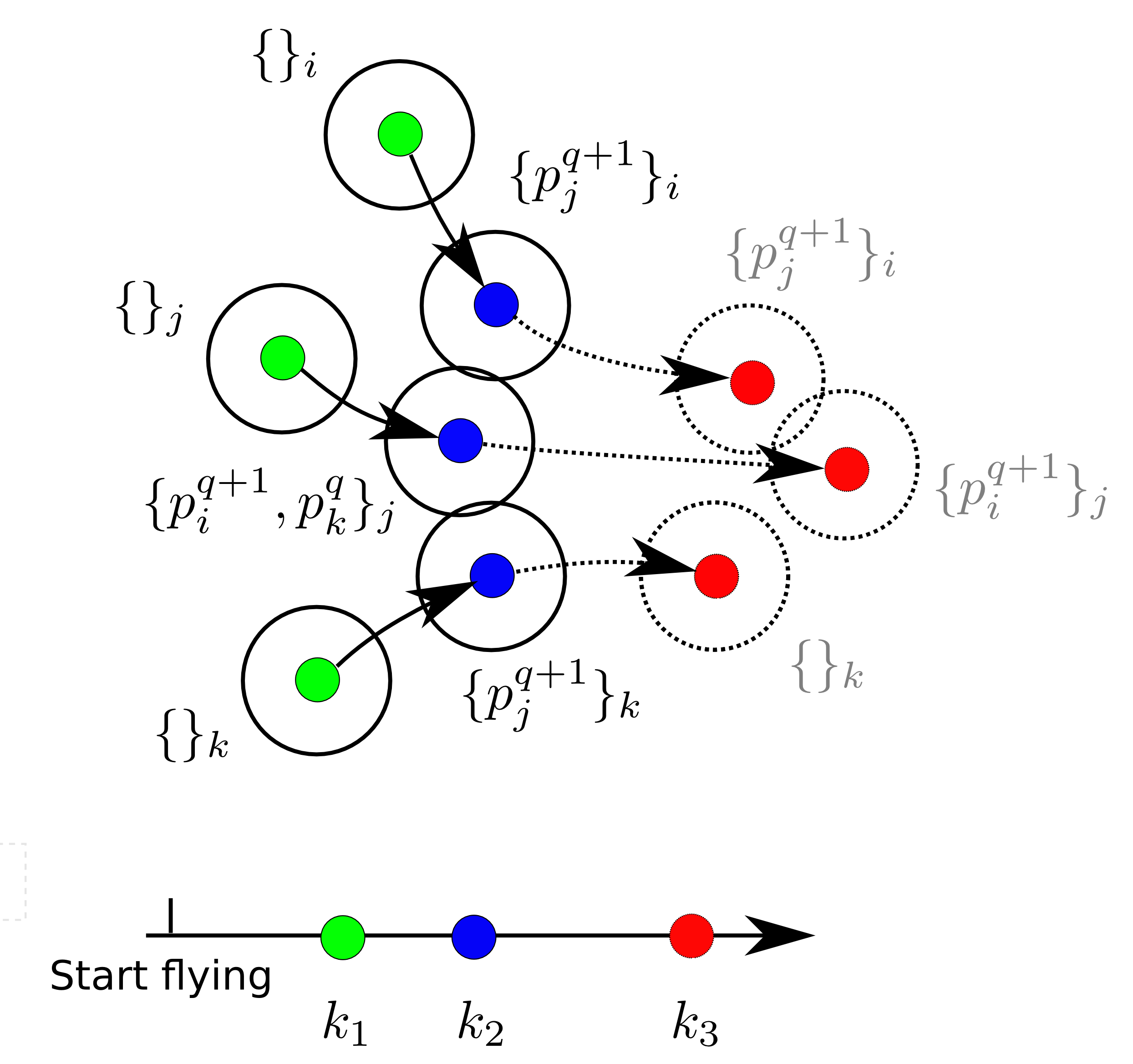}
	\caption{At $k_2$, $i,j$ re-optimize and share the collision-free trajectories to their neighbors}	
	\label{fig:afterOpt}	
	\end{subfigure}
\caption{Illustration of Alg.~\ref{alg:dis_tra_opt}}
\label{fig:algorithm}
\end{figure}

To address above problems, we first make three key observations:
\begin{enumerate}
\item For the distributed projected algorithm~\ref{alg:dis_proj_alg}, each Crazyflie optimizes the trajectories of the remaining $N-1$ Crazyflies. Let $x_i=[x^T_{i1},...,x^T_{iN}]^T$, where $x_{ij}\in \Re^{3K}$ is Crazyflie $i$'s solution of the trajectory of Crazyflie $j$. We may limit Crazyflie $i$ to optimize only $x_{ii}$, to accept other Crazyflies' optimized trajectories $x_{ij}$ directly without the averaging step, and to treat $x_{ij}$ of other Crazyflies as static obstacles. Because the trajectories of other Crazyflies will not be optimized, Eqn.(\ref{eq:collisionSinglei})-(\ref{eq:collisionSinglej}) are used for the convexification and the initial-to-final state constraints of other Crazyflies are also excluded from $X_i$.  As a result, the distributed projected algorithm becomes:
\begin{align}
x_{ii}(m+1) &= P_{\bar{X}_{i}}\left[x_{ii}(m)\right], \quad \bar{X}_i=\left\{\begin{aligned}A_{eq,i}\ x_{ii}=b_{eq,i}\\ \bar{A}_{in,i}\ x_{ii} \preceq \bar{b}_{in,i} \end{aligned}\right. \\
x_{ij}(m+1) &= x_{jj}(m),\quad j\in N_i \label{eq:x_ij=x_jj}\\ 
x_{ik}(m+1) &= x_{jk}(m),\quad k\notin N_i,\ j\in N_i \text{ and if $x_{jk}$ is the most updated copy of $x_{kk}$}
\label{eq:x_ik=x_jk}
\end{align}
where $A_{eq,i}\ x_{ii}=b_{eq,i}$ is the initial-to-final state constraint of Crazyflie $i$ and $\bar{A}_{in,i}\ x_{ii} \preceq \bar{b}_{in,i}$ has the collision constraints convexified using Eqn.~\ref{eq:collisionSinglei})-~\ref{eq:collisionSinglej} and other constraints local to $i$. Note that both $X_i$ and $\bar{X}_i$ have the same number of constraints but with different dimensionality. Eqn.~\ref{eq:x_ij=x_jj} means that Crazyflie $i$'s solution of neighbor $j$ is updated with the solution Crazyflie $j$ has optimized itself. Eqn.~\ref{eq:x_ik=x_jk} means that to update non-neighbors' solution $x_{ik}$, Crazyflie $i$ selects the most updated one among the neighbors' solutions. 

\item Due to the presence of collision avoidance constraints, along the resulting optimized trajectories, Crazyflies are separated by a significant amount of space from their neighbors. For those Crazyflies who are not neighbors, they are separated by other Crazyflies in between. Therefore the collision avoidance constraints for non-neighbor pairs are actually not active after all. We then could exclude these inactive constraints pairs to reduce the optimization time. This is achieved by defining an collision active region with radius of $R_{active}$ such that only those Crazyflies who are within this region are neighbors with active collision avoidance constraints. As shown in Fig.~\ref{fig:algorithm}, there are no neighbors detected at time $k_1$, whereas at time $k_2$ Crazyflie $i,j,k$ detected neighbors among themselves because they are sufficiently close to each other.

\item If we are able to exclude inactive constraints for non-neighbors, then constraints involving $x_{ik},\ k\notin N_i$ are excluded from $\bar{A}_{in,i}\ x\preceq \bar{b}_{in,i}$. Crazyflie $i$ then only need to accept neighbors' solution $x_{ij},\ j\in N_i$. The algorithm becomes:
\begin{align}
x_{ii}(m+1) &= P_{\tilde{X}_{i}}\left[x_{ii}(m)\right], \quad \tilde{X}_i=\left\{\begin{aligned}A_{eq,i}\ x_{ii}&=b_{eq,i}\\ \tilde{A}_{in,i}\ x_{ii} &\preceq \tilde{b}_{in,i},\ \text{no non-neighbor constraints} \end{aligned}\right. \label{eq:x_ii=P_Xi_hat} \\
x_{ij}(m+1) &= x_{jj}(m),\quad \forall j\in N_i \label{eq:x_ij=x_jjobs3}
\end{align}
\end{enumerate}

\begin{remark}
Without collision constraints with non-neighbors, the optimization time is significantly reduced and Crazyflie $j$ does not need to communicate $x_{jk}$ to Crazyflie $i$ any more. Therefore the communication costs are also lowered. 
\end{remark}
\begin{algorithm}
	\caption{Distributed trajectory optimization}
	\label{alg:dis_tra_opt}
	\begin{algorithmic}[1]
		\For{\textbf{each} Crazyflie $i$} 
		\State $(p_{ii},v_{ii},x_{ii})\leftarrow$  straightLine$(p_{i}[0],v_{i}[0],p_{i}[KT],v_{i}[KT])$
		\State $k_{c,i}\leftarrow\underset{k\in 1,...,K}{\text{argmin}}\|p_i[k]-r_o\|^2_2$
		\State $(p_i,v_i,a_i)\leftarrow$  crossingCenter$(p_i[0],v_i[0],p_i[KT],v_i[KT],p_i[k_{c,i}T],v_i[k_{c,i}T])$
		\EndFor
		\For{\textbf{each} $k=1,...,K$}
			\State $m\leftarrow 0$
			\For{\textbf{all} Crazyflie $i$}
				\State obstacleSet(i) $\leftarrow$ $\{p_{ij}\ |\ \|p_{ii}[k]-p_{ij}[k]\|_2\leq R_{active}, \ \forall j\in 1,...,N,\ j\neq i\}$ 
				\While {existCollision($p_{ii}$, obstacleSet(i)) \textbf{and} $m<M_1$}					
						\State $\tilde{A}_{in,i}\ x_{ii}\preceq \tilde{b}_{in,i}\leftarrow \text{2ndConvexification}(x_{ii},\ obstacleSet(i))$
						\State $x_{ii} \leftarrow P_{\tilde{X}_i}[x_{ii}]$ \label{line:P_X}
						\State $p_{ii} \leftarrow x_{ii}$
						\For {\textbf{all} $j\in N_i$}
						\State $x_{ij}\leftarrow x_{jj}$
						\State $p_{ij}\leftarrow x_{ij}$ 										\State $\text{obstacleSet(i)} \leftarrow p_{ij}$
						\EndFor
						\State $m\leftarrow m+1$
				\EndWhile
				\State Crazyflie $i$ tracks $(p_{ii}[k],v_{ii}[k],x_{ii}[k])$			
			\EndFor
		\EndFor
	\end{algorithmic}
\end{algorithm}

\subsection{The Algorithm}
We summarize above observations in Alg.~\ref{alg:dis_tra_opt}. Fig.~\ref{fig:algorithm} is an example that illustrates the process of running Alg.~\ref{alg:dis_tra_opt}. At the beginning the flight, each Crazyflie computes an initial trajectory passing through the center of ring as explained in section~\ref{sec:initialSolution} and start tracking the initial trajectory. At each time instant it is detecting neighbors who are within a sphere of radius $R_{active}$ around it and add their trajectories to its obstacle set. The obstacle sets are used for convexifications of both collision avoidance constraints (Eqn.~\ref{eq:collisionSinglei}- ~\ref{eq:collisionSinglej}) and the ring constraints (Eqn.~\ref{eq:tube}-~\ref{eq:rightCone}). If there will be collisions between its nominal trajectory $x_{ii}$ and neighbors' trajectories $x_{ij}$ in future states, $x_{ii}$ will be re-optimized with Eqn.~\ref{eq:x_ii=P_Xi_hat} such that future collisions are avoided and the \textit{new} trajectory is shared to their neighbors (Eqn.~\ref{eq:x_ij=x_jjobs3}). In Fig.~\ref{fig:afterOpt} Crazyflie $i,j$ re-optimize and share their trajectories to their neighbors at $k_2$, whereas Crazyflie $k$ does not optimize its trajectory since there are no future collision detected with $j$. Crazyflie $k$ only updates the obstacle set from $\{p^q_j\}_k$ to $\{p^{q+1}_j\}_k$ after having received the re-optimized trajectory $p^{q+1}_j$. Note that because Crazyflie $k$ does not re-optimize its trajectory, the obstacle set of Crazyflie $j$ keeps $p^q_k$ unchanged at $k_2$. 

Alg.~\ref{alg:dis_tra_opt} allows each Crazyflie to do trajectory optimization in parallel while they are flying. Their neighbors' trajectories are dynamically added to or removed from the obstacle sets depending on the closeness to their neighbors. Thus the number of constraints and the solving time of each optimization for each Crazyflie is limited as much as possible. Unlike model predictive control, Alg.\ref{alg:dis_tra_opt} solves optimization only when collisions are detected in the nominal trajectories, but otherwise Crazyflies will only follow the nominal trajectories without recomputing new trajectories. In a word, the trajectory optimization of the swarm was solved in parallel by each Crazyflie with minimal number of collision constraints at only several time instants when collisions were detected.

\begin{figure}
\centering
	\begin{subfigure}[h]{0.3\textwidth}
		\includegraphics[width=\textwidth]{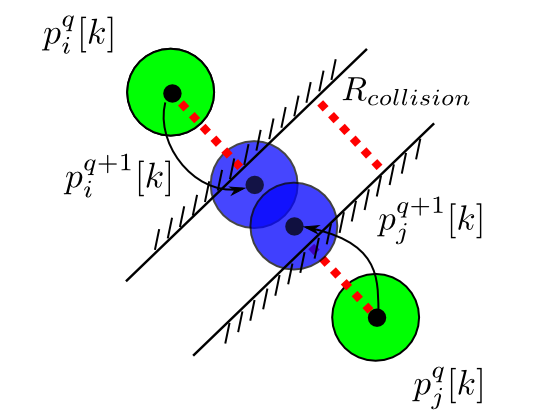}
		\caption{$p^{q+1}_i[k]$ and $p^{q+1}_j[k]$ are not collision-free}
		\label{fig:compare_collision_left}
	\end{subfigure}
	~
	\begin{subfigure}[h]{0.3\textwidth}
		\includegraphics[width=\textwidth]{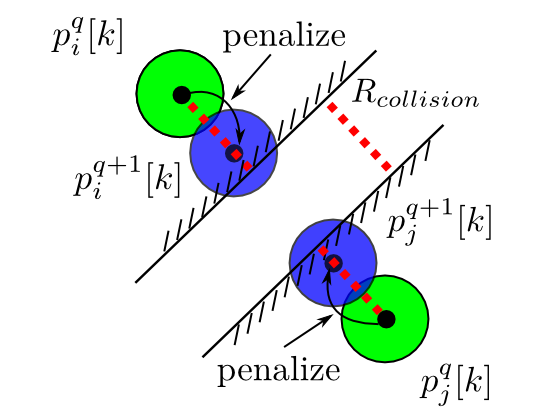}
		\caption{Penalize deviation to reduce the possibility of collision}
		\label{fig:compare_collision_middle}
	\end{subfigure}
	\begin{subfigure}[h]{0.3\textwidth}
		\includegraphics[width=\textwidth]{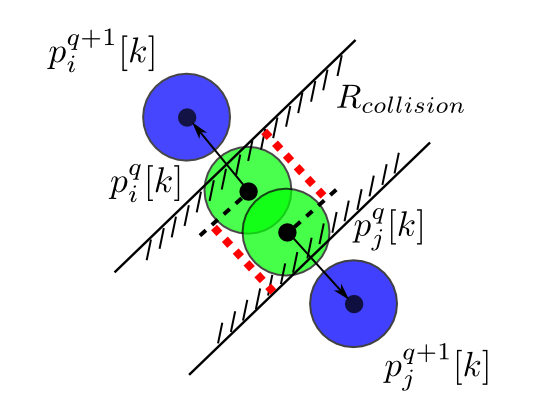}
		\caption{Violated constraint always results in collision-free positions}
		\label{fig:compare_collision_right}
	\end{subfigure}
	\caption{Illustration of projection operation}
	\label{fig:compare_collision}
	~
	\begin{subfigure}[h]{0.45\textwidth}
		\includegraphics[width=\textwidth]{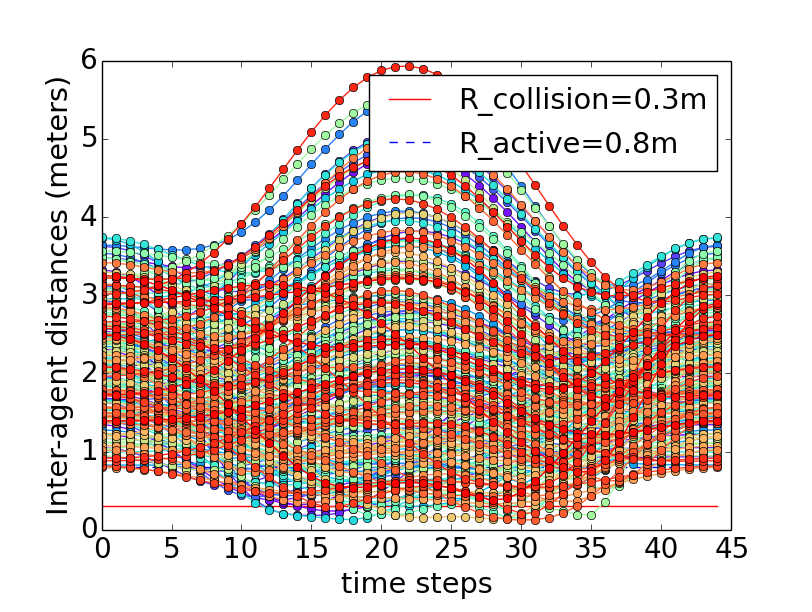}
		\caption{Initial solutions}
		\label{fig:distance_initial}
	\end{subfigure}
	\begin{subfigure}[h]{0.45\textwidth}
		\includegraphics[width=\textwidth]{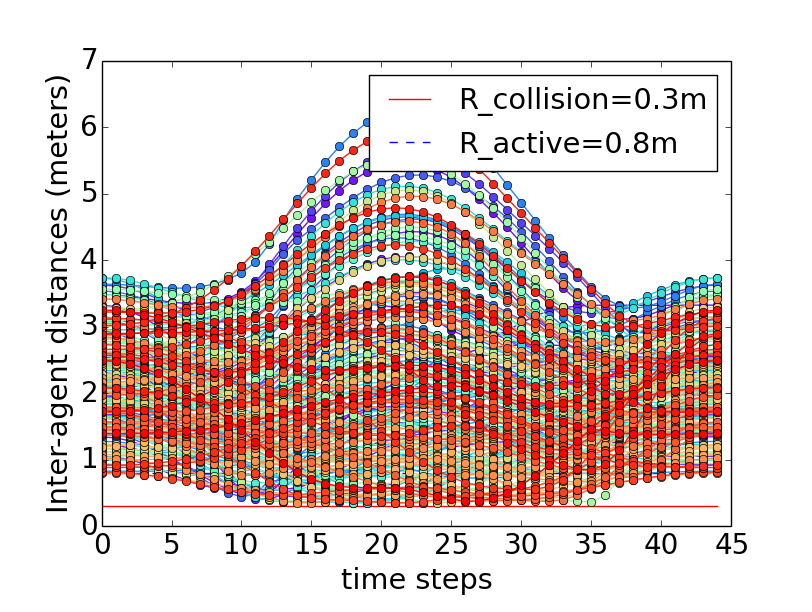}
		\caption{Final solutions}
		\label{fig:distance_final}
	\end{subfigure}
	\caption{Inter-Crazyflie distances of 20 Crazyflies}
	\label{fig:distances}
	~
	\includegraphics[width=0.6\textwidth]{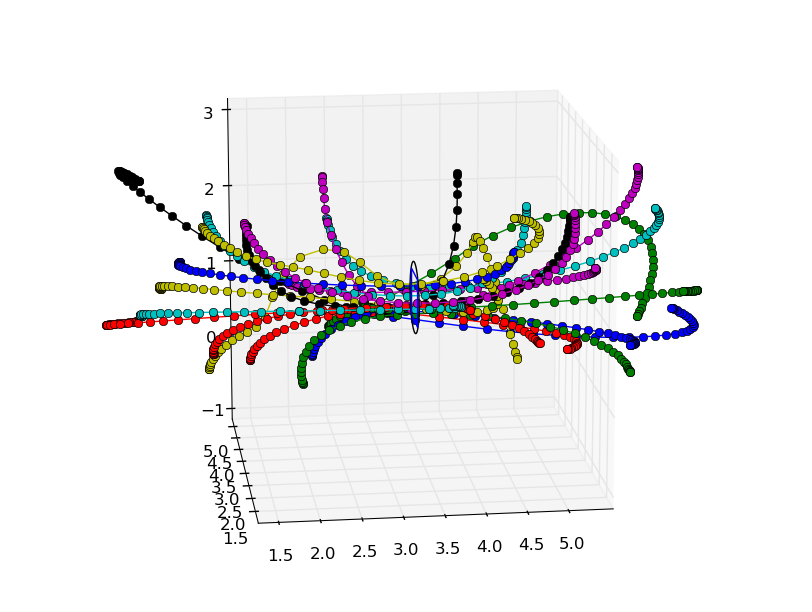}
	\caption{Planned trajectories through an opening from left to right for 20 Crazyflies}
	\label{fig:20_agents}
\end{figure}

\subsection{Convergence}
As shown in Fig.~\ref{fig:compare_collision_left}, since Crazyflie $i,j$ optimize trajectories independently with respect to $p^q_j,p^q_i$, the resulting re-optimized trajectories $p^{q+1}_i,p^{q+1}_j$ are only guaranteed to be collision-free with respect to $p^q_j,p^q_i$ but may not be collision-free between themselves. Crazyflie $i,j$ need another optimization if there exist collisions between $p^{q+1}_j$ and $p^{q+1}_i$ and repeat the optimization until the trajectories are collision-free (assume at most $M_1$ repetitions). Strictly speaking the convergence is not guaranteed~\cite{morgan2014model}. Nonetheless, Alg.~\ref{alg:dis_tra_opt} reduces the possibility of convergence failure by solving a projection problem in line~\ref{line:P_X}, where the objective of projection problem penalizes the deviation of optimized solution to previous solution. Therefore positions that are already collision-free in previous solutions will be preserved to be collision-free as much as possible. This is illustrated in Fig.~\ref{fig:compare_collision_middle}. Note that in Fig.~\ref{fig:compare_collision_right} collision constraint convexified from collision violated positions will guarantee the re-optimized positions are collision-free. Although solving projection problem also does not guarantee convergence, experiments show that convergence failures seldomly happen. An example of optimized trajectories for 20 Crazyflies are illustrated in Fig.~\ref{fig:20_agents} and the inter-Crazyflie distances of the initial solutions as well as optimized solutions are shown in Fig.~\ref{fig:distances}. The radius of the ring is set to $R_{ring}=0.6$m. The duration of flight is $T=Kh=40\times 0.15s=6s$. The solver we used is ECOS, which is efficient and open-source~\cite{Domahidi2013ecos}.
\clearpage
\subsection{Result Comparison}
\begin{figure}[h]
\centering
	\begin{subfigure}[h]{0.42\textwidth}
		\includegraphics[width=\textwidth]{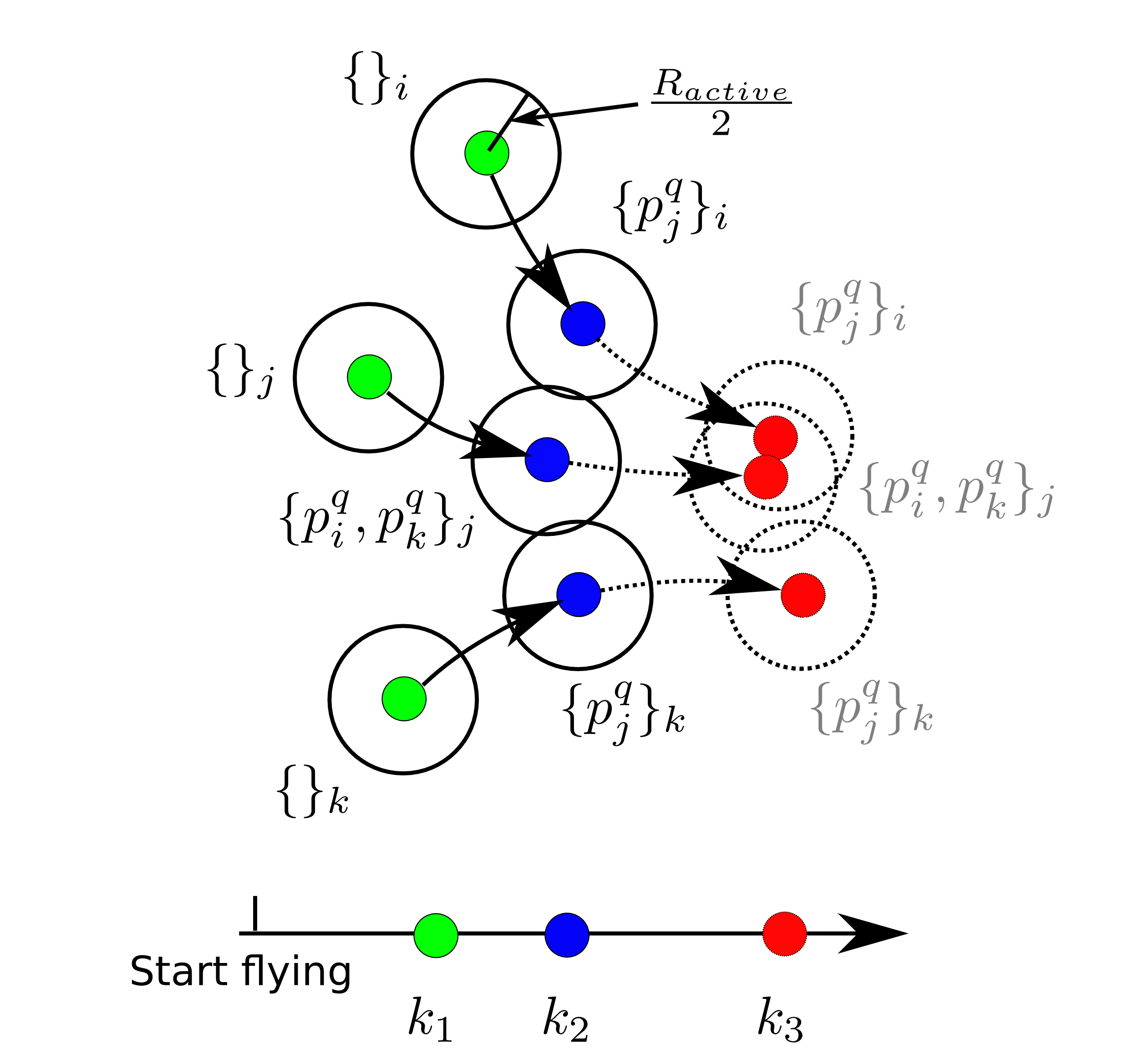}
		\caption{Alg.\ref{alg:dis_tra_opt} dynamically adds or removes constraints}
		\label{fig:compare_left}
	\end{subfigure}
	~
	\begin{subfigure}[h]{0.5\textwidth}
		\includegraphics[width=\textwidth]{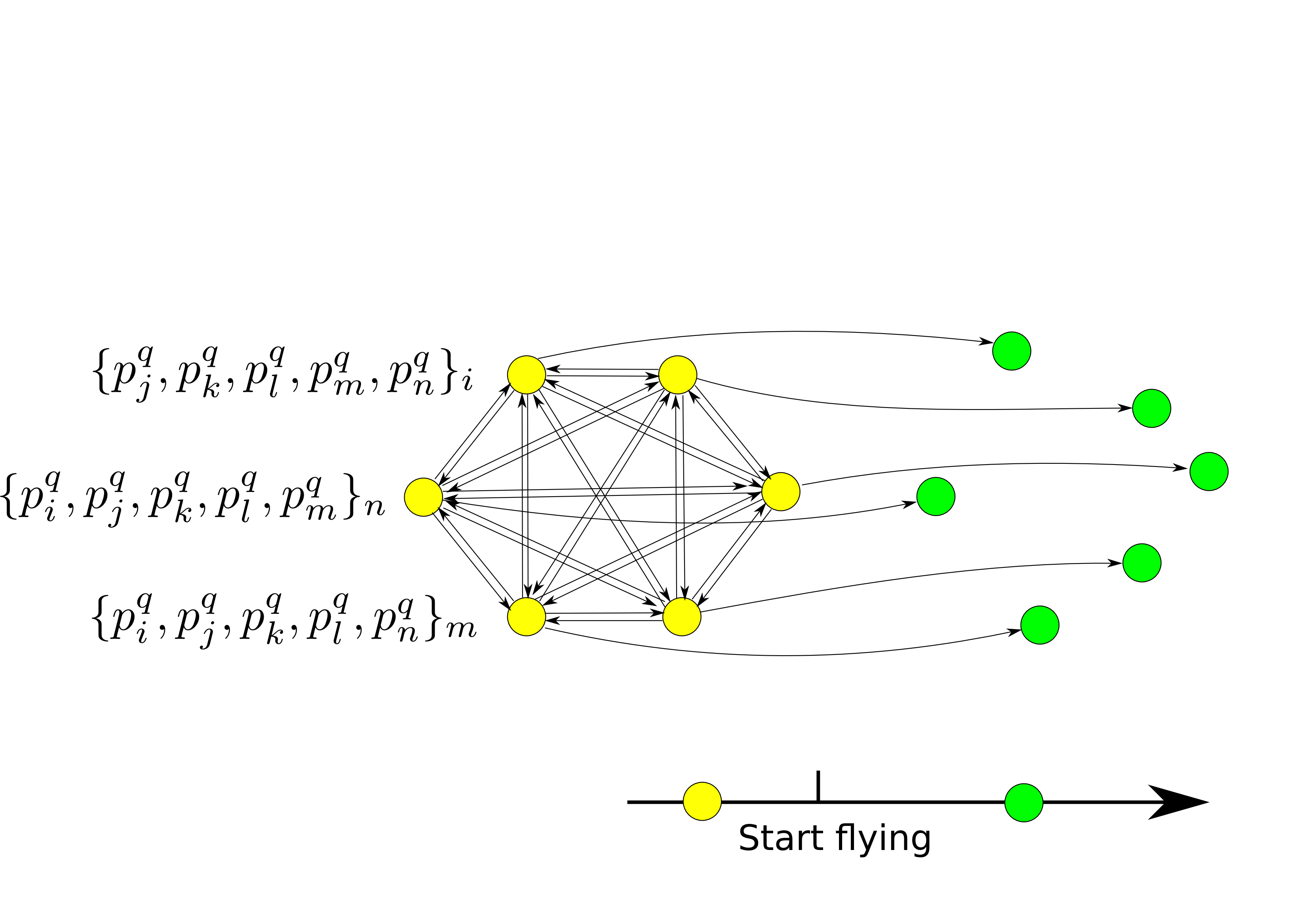}
		\caption{All collision avoidance constraints with other nodes are taken into account}
		\label{fig:compare_right}
	\end{subfigure}
\caption{Compare Alg.\ref{alg:dis_tra_opt} with a decentralized approach}
\label{fig:compare_diagram}
	~
	\begin{subfigure}[h]{0.42\textwidth}
		\includegraphics[width=\textwidth]{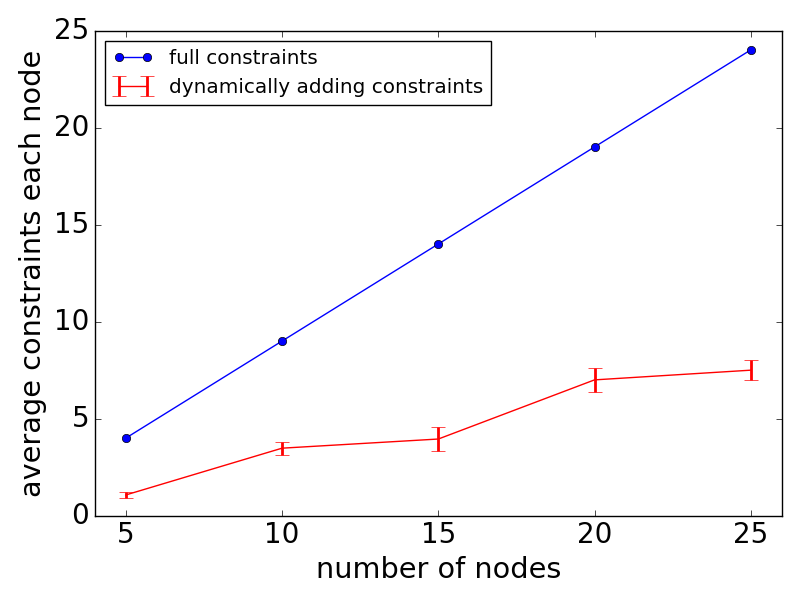}
		\caption{Average constraints per node v.s. number of nodes}
	\end{subfigure}
	~
		\begin{subfigure}[h]{0.42\textwidth}
		\includegraphics[width=\textwidth]{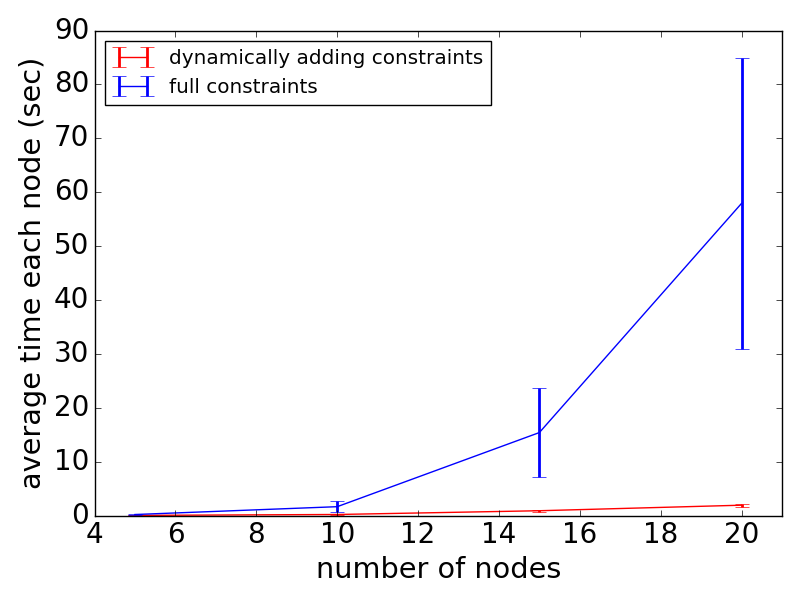}
		\caption{Average optimization time per node v.s. number of nodes}
	\end{subfigure}
\caption{Performance comparison}
\label{fig:performance_comparison}
\end{figure}

In this section we compare the performance of Alg.~\ref{alg:dis_tra_opt} with a decentralized approach that does not define a collision active region for dynamically adding or removing constraints. The decentralized approach will solve the optimization before the flight and each Crazyflie/node includes all collision avoidance constraints with others. As shown in Fig.~\ref{fig:compare_right}, the collision set of each node contains all the trajectories of other nodes.  

We solved both trajectory optimizations for 20 times and averaged the results. We compared the performance of these two approaches according to:
\begin{enumerate}
\item The average number of collision constraints for each Crazyflie
\item The average solving time for each Crazyflie
\end{enumerate}

Fig.~\ref{fig:performance_comparison} demonstrates that the average number of constraints $\bar{N}_{collision}$ approaches  $\bar{N}_{neighbor}\approx 8$ trajectories for Alg.~\ref{alg:dis_tra_opt} whereas for the decentralized approach it grows linearly. In addition, the average solving time of Alg.~\ref{alg:dis_tra_opt} is much less than the decentralized approach, which is explained in remark~\ref{rm:avg_time_left} and \ref{rm:avg_time_right}, and the runtime of Alg.~\ref{alg:dis_tra_opt} whenever a collision detected is of order $O(M_1^2\bar{N}^2_{neighbors})$.

\begin{remark}\label{rm:avg_time_left}
Alg.~\ref{alg:dis_tra_opt} solves optimization whenever collisions are detected during the flight. Let $M_1$ be the number of such optimizations, and the average solving time for Alg.~\ref{alg:dis_tra_opt} is defined as $T_{avg,1}=\frac{T_{total,1}}{M_1N}$, whereas the decentralized approach is $T_{avg,2}=\frac{T_{total,2}}{N}$. Because of the division of $M_1$, the average solving time of Alg.~\ref{alg:dis_tra_opt} is much less the decentralized one. The reason for dividing $M_1$ is that for real time application, Crazyflie will track the optimized trajectory as soon as each optimization completes. Thus the solving time for each optimization is more interesting than for the total time for all optimizations each Crazyflie has ever solved.
\end{remark}

\begin{remark}\label{rm:avg_time_right}
The decentralized approach in Fig.~\ref{fig:compare_right} often fail to find feasible solutions during optimization and need multiple times of re-convexification and re-optimization before finding the feasible solution. This is one main reason why it takes much longer time than the Alg.~\ref{alg:dis_tra_opt}. The frequent optimization failure may be due to too many convexified non-neighbor constraints were included and the feasible region became too small. Note that the re-convexification is done by the convexification of the \textit{infeasible} solution returned by the solver.
\end{remark}

\clearpage
\section{Future Improvement}
Alg.~\ref{alg:dis_tra_opt} is fast because the re-optimized trajectory will be directed accepted by neighbors. However Alg.~\ref{alg:dis_tra_opt} convexifies the collision avoidance constraints using Eqn.~\ref{eq:collisionSinglei}-Eqn.~\ref{eq:collisionSinglej} (2ndConvexification), which significantly limits the feasible region of the optimization problem. Besides, the convergence is not guaranteed using Eqn.~\ref{eq:collisionSinglei}-Eqn.~\ref{eq:collisionSinglej}, even though the projection operation may reduce this possibility. Here we propose an improvement on the Alg.~\ref{alg:dis_tra_opt} that uses the convexification Eqn.~\ref{eq:collisionDouble} (1stConvexification) so that convergence is not an issue any more at the expense of more frequent communications and optimizations. The idea is that Crazyflie $i$ optimize both its own and neighbors' trajectories when collisions are detected. The algorithm is shown in Alg.~\ref{alg:dis_tra_opt_2}. Similar to Alg.~\ref{alg:dis_proj_alg}, Crazyflies that are running Alg.~\ref{alg:dis_tra_opt_2} average the neighbors' and their own solutions to reach consensus, and repeatedly project the averaged solution to local feasible sets. We assume by at most $M_2$ communications the consensus can be reached. The difference is that for Alg.~\ref{alg:dis_tra_opt_2}, Crazyflies will not optimize non-neighbors solutions. Since the average number of neighbors approaches a limit, the number of neighbors' trajectories to optimize is also limited. Therefore Alg.~\ref{alg:dis_tra_opt_2} is also scalable with the number of Crazyflies. Given communication time of the trajectories and runtime of projection are sufficiently small, Alg.~\ref{alg:dis_tra_opt_2} is superior to Alg.~\ref{alg:dis_tra_opt} because the convergence issue due to convexification does not exist any more and the convexified constraints have larger feasible regions. The runtime of Alg.~\ref{alg:dis_tra_opt_2} is of order $O(M^2_2\bar{N}^2_{neighbors})$.

\begin{algorithm}
	\caption{Distributed trajectory optimization}
	\label{alg:dis_tra_opt_2}
	\begin{algorithmic}[1]
		\For{\textbf{each} Crazyflie $i$} 
		\State $(p_{ii},v_{ii},x_{ii})\leftarrow$  straightLine$(p_{i}[0],v_{i}[0],p_{i}[KT],v_{i}[KT])$
		\State $k_{c,i}\leftarrow\underset{k\in 1,...,K}{\text{argmin}}\|p_i[k]-r_o\|^2_2$
		\State $(p_i,v_i,a_i)\leftarrow$  crossingCenter$(p_i[0],v_i[0],p_i[KT],v_i[KT],p_i[k_{c,i}T],v_i[k_{c,i}T])$
		\EndFor
		\For{\textbf{each} $k=1,...,K$}
			\State $m\leftarrow 0$
			\For{\textbf{all} Crazyflie $i$}
				\State obstacleSet(i) $\leftarrow$ $\{p_{ij}\ |\ \|p_{ii}[k]-p_{ij}[k]\|_2\leq R_{active}, \ \forall j\in 1,...,N,\ j\neq i\}$ 
				\If {existCollision($p_{ii}$, obstacleSet(i))}					
						\State $\tilde{A}_{in,i}\ [x^T_{ii},x^T_{ij},...]^T\preceq \tilde{b}_{in,i}\leftarrow \text{1stConvexification}(x_{ii},\ obstacleSet(i))$
						\State $[x^T_{ii},x^T_{ij},...]^T \leftarrow P_{\tilde{X}_i}\left[[x^T_{ii},x^T_{ij},...]^T\right]$
						\While {$\|x_{ij}(m+1)-x_{ij}(m)\|_2>\epsilon$ \textbf{and} $m<M_2$} \Comment{Test convergence}
						\For {\textbf{each} $j\in N_i\cup i$}
						\State $\begin{aligned}&x_{ij}\leftarrow \sum_{s\in N_i\cup i}\frac{1}{N_i+1} x_{sj}\end{aligned}$
						\State $p_{ij}\leftarrow x_{ij}$
						\State $\text{obstacleSet(i)} \leftarrow p_{ij}$ \Comment{Update collision set}
						\EndFor	
						\State $m\leftarrow m+1$					
						\EndWhile						
				\EndIf
				\State Crazyflie $i$ tracks $(p_{ii}[k],v_{ii}[k],x_{ii}[k])$			
			\EndFor
		\EndFor
	\end{algorithmic}
\end{algorithm}

\cleardoublepage
\chapter*{Conclusion}
\addcontentsline{toc}{chapter}{Conclusion}
This master thesis presented a distributed system to enable a swarm of quadcopters to fly through the openings. The distributed estimation, control and optimization techniques were discussed in details to achieve the goal of the project: the quadcopter swarm is able to fly through an opening subject to local communication and measurement constraints. 

We demonstrated that the bearing and distance sensors can be used for localization given one Crazyflie has global position measurement, and that the coupled linear and rotational dynamics of quadcopters that allows Crazyflies to estimate their attitude is crucial to the localization. Since the majority of existing work about distributed control and estimation assumed point mass model without exploring the real dynamics of agent, our work could motivate people to take advantage of the dynamics of agents in future. 

We also drew a conclusion that the distributed control is only suitable in an environment of free space. For complex environment, trajectory optimization is necessary to accomplish the challenging tasks that are often difficult for distributed control such as collision avoidance. We had presented the procedures of adapting a distributed optimization method to two trajectory optimization algorithms, and discussed the performance of the first algorithm Alg.~\ref{alg:dis_tra_opt}. The thesis was finalized with the future work where the second scalable trajectory optimization algorithm Alg.~\ref{alg:dis_tra_opt_2} was proposed and comparisons to Alg.~\ref{alg:dis_tra_opt} were highlighted.

\section*{Acknowledgement}
Here I would like to express my sincere gratitude to Michael Hamer for giving me an opportunity to carry out the master project in Professor D'Andrea's group. I also thank for his support, patience, continuous guidance and insightful suggestions in the past 6 months. He has been always playing a key role of encouraging me and keeping me on the right track during the course of this project.
\cleardoublepage
% ...



\bibliographystyle{plain}
\addcontentsline{toc}{chapter}{Bibliography}
\bibliography{bibliography}

\end{document}